\documentclass[preprint,12pt]{elsarticle}

\usepackage{amssymb}
\usepackage{amsmath}
\usepackage{graphicx}   % 用于插入图片
\usepackage{booktabs}   % 用于三线表
\usepackage{multirow}   % 用于表格合并行
\usepackage{caption}    % 用于图表标题设置
\usepackage{hyperref}   % 用于交叉引用和超链接
\usepackage{float}      % 用于强制图片位置 [H]
\usepackage{lineno}     % 用于生成行号(审稿通常需要)

\usepackage{xcolor}
\usepackage{colortbl}
\usepackage{array}      % 用于表格格式控制
\usepackage{tabularx}   % 用于自适应表格宽度

\journal{Medical Image Analysis}

\begin{document}

\begin{frontmatter}

%% ===============================================
%% Title
%% ===============================================
\title{Physically-Grounded Manifold Projection Model for Generalizable Metal Artifact Reduction in Dental CBCT}

%% ===============================================
%% Authors & Affiliations
%% ===============================================
%% --- Authors ---

%% 1. Zhi Li (First Author, Equal Contribution)
\author[1]{Zhi Li\fnref{fn1}}
% \ead{zhi.li@hdu.edu.cn}

%% 2. Yaqi Wang (Co-First Author, Equal Contribution)
\author[2]{Yaqi Wang\fnref{fn1}}
% \ead{wangyaqi@cuz.edu.cn}

%% 3. Bingtao Ma (Added after Yaqi Wang)
\author[1]{Bingtao Ma}
% \ead{ma@example.com}

%% 4. Yifan Zhang
\author[3]{Yifan Zhang}
% \ead{zhang@example.com}

%% 5. Huiyu Zhou (Co-Corresponding Author)
\author[4]{Huiyu Zhou\corref{cor1}}
% \ead{hz143@leicester.ac.uk}

%% 6. Shuai Wang (First Corresponding Author)
\author[1]{Shuai Wang\corref{cor1}}

%% --- Footnotes for Roles ---
\cortext[cor1]{Corresponding authors. \ead{shuaiwang.tai@gmail.com}} 
\fntext[fn1]{These authors contributed equally to this work.}

%% --- Affiliations ---

%% Affiliation 1: Hangzhou Dianzi University (School of Cyberspace)
\affiliation[1]{organization={School of Cyberspace, Hangzhou Dianzi University},
                city={Hangzhou},
                postcode={310018},
                country={China}}

%% Affiliation 2: Hangzhou Dianzi University (Innovation Center)
\affiliation[2]{organization={Innovation Center for Electronic Design Automation Technology, Hangzhou Dianzi University},
                city={Hangzhou},
                postcode={310018},
                country={China}}

%% Affiliation 3: Hangzhou Geriatric Stomatology Hospital
\affiliation[3]{organization={Hangzhou Geriatric Stomatology Hospital, Hangzhou Dental Hospital Group},
                city={Hangzhou},
                country={China}}

%% Affiliation 4: University of Leicester
\affiliation[4]{organization={School of Computing and Mathematical Sciences, University of Leicester},
                city={Leicester},
                postcode={LE1 7RH},
                country={United Kingdom}}

% 
%% ===============================================
%% Abstract
%% ===============================================

\begin{abstract}
Metal artifacts in Dental Cone-Beam Computed Tomography (CBCT) pose a significant impediment to diagnostic precision by severely obscuring anatomical structures. While deep learning has propelled the progress of Metal Artifact Reduction (MAR), current paradigms encounter intrinsic limitations: supervised regression methods often suffer from the ``regression-to-the-mean'' phenomenon, leading to spectral blurring, while unsupervised approaches lack pixel-wise constraints, inducing structural hallucinations. Recently, Denoising Diffusion Probabilistic Models (DDPMs) have offered superior texture realism but rely on computationally intensive iterative sampling ($\epsilon$-prediction), which introduces stochastic uncertainty compatible with art generation but risky for medical diagnosis. To reconcile these methodological conflicts, we propose the Physically-Grounded Manifold Projection (PGMP) framework. First, addressing the data bottleneck, we establish an Anatomically-Adaptive Physics Simulation (AAPS) pipeline. By synergizing Monte Carlo-inspired polychromatic spectral modeling with patient-specific digital twins, this module synthesizes high-fidelity training pairs that effectively bridge the synthetic-to-real domain gap. Second, to overcome the inefficiency of stochastic sampling, we strategically adapt the Direct $x$-Prediction paradigm via our DMP-Former. Grounded in the Manifold Assumption, this module reformulates restoration as a deterministic projection mapping, recovering clean anatomical structures in a single forward pass. Finally, to guarantee clinical plausibility, we incorporate a Semantic-Structural Alignment (SSA) mechanism. This module distills domain-specific priors from medical foundation models (MedDINOv3), acting as a digital anatomist to anchor the solution space and prevent hallucinatory artifacts. Extensive experiments across synthetic and multi-center clinical datasets demonstrate that PGMP not only outperforms state-of-the-art baselines on unseen anatomy but also sets a new benchmark for computational efficiency and diagnostic reliability. Our code and dataset will be publicly available at \url{https://github.com/ricoleehduu/PGMP}.
\end{abstract}

%% ===============================================
%% Keywords
%% ===============================================

\end{frontmatter}

%% 开启行号（审稿模式通常建议开启）
% \linenumbers

%% ===============================================
%% Main Content Placeholder
%% ===============================================
% \section{Introduction}
% ...
%% ===============================================
%% Section 1: Introduction
%% ===============================================
\section{Introduction}
\label{sec:intro}

Cone-Beam Computed Tomography (CBCT) stands as the cornerstone of modern digital dentistry, providing indispensable 3D volumetric visualization for implantology, orthodontics, and pathology diagnosis \cite{kapila2015cbct, weiss2019cone}. However, the inevitable presence of high-density metallic implants (e.g., titanium fixtures, amalgam fillings) frequently introduces severe metal artifacts (MA). Manifesting as dark bands (photon starvation) and bright streaks (beam hardening), these artifacts not only degrade visual quality \cite{klotz1987reduction, zhang2018convolutional} but, more critically, obliterate the structural integrity of adjacent tissues. This degradation severely limits diagnostic confidence and hampers the efficacy of downstream computerized tasks \cite{song2009correlation, shan2020abnormal}, making effective Metal Artifact Reduction (MAR) a prerequisite for reliable clinical decision-making.

The methodology for MAR has evolved through a pendulum swing between determinism and stochasticity. In the past decade, deep learning revolutionized the field, with early supervised approaches formulating MAR as a deterministic regression task \cite{zhang2018convolutional, lin2019dudonet}. While computationally efficient, these pixel-wise regression models often suffer from the ``regression to the mean'' phenomenon, producing over-smoothed, waxy textures that fail to preserve fine trabecular bone details \cite{wang2023semimar}. Furthermore, their reliance on paired training data creates a substantial synthetic-to-real domain gap, as conventional monochromatic simulations fail to replicate the complex polychromatic scattering of clinical acquisitions \cite{liu2024unsupervised}. Conversely, the recent wave of Generative Diffusion Models (DDPMs) has achieved unprecedented texture realism by modeling the data distribution via iterative noise removal \cite{karageorgos2024denoising, zhang2025coupled}. However, this generative prowess comes with critical bottlenecks: the computationally intensive iterative sampling ($T \gg 100$) prohibits real-time clinical workflows, and the inherent stochasticity introduces aleatoric uncertainty, leading to the risk of \textit{structural hallucinations} where the model invents plausible but non-existent anatomy \cite{ma2025radiologist}.

%% Insertion of Figure 1
\begin{figure}[t]
	\centering
	\includegraphics[width=1.0\linewidth]{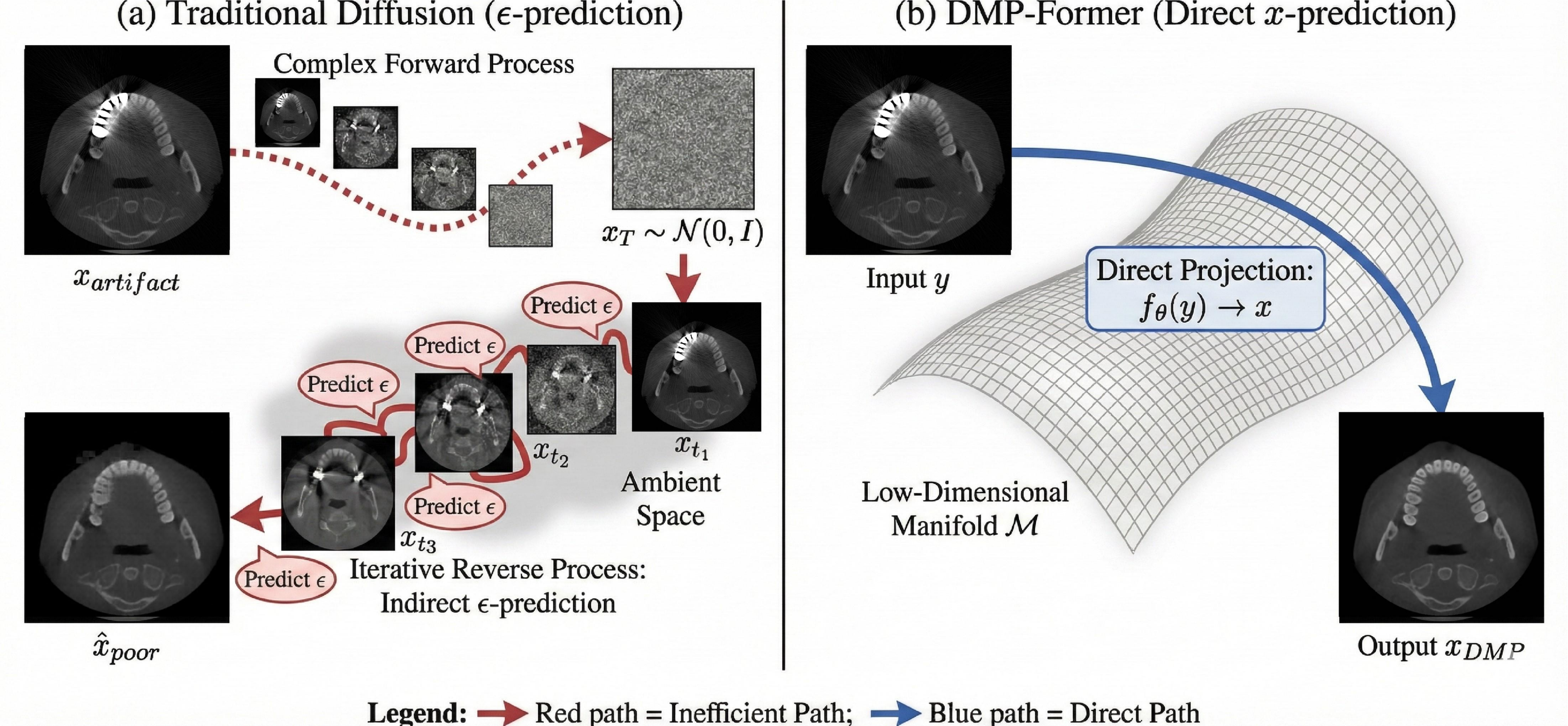}
	\caption{Conceptual comparison between the stochastic noise-prediction paradigm and our deterministic manifold projection. (a) Traditional Diffusion ($\epsilon$-prediction): Iteratively removes Gaussian noise $\epsilon \sim \mathcal{N}(0, I)$ by estimating the score function in the high-dimensional ambient space. This process is computationally intensive ($T \gg 1$) and prone to accumulating stochastic errors. (b) Proposed DMP-Former ($x$-prediction): Drawing inspiration from Consistency Models, we learn a direct projection mapping $f_\theta: Y \rightarrow \mathcal{M}$. This paradigm bypasses the chaotic noise trajectory, ensuring structural determinism and efficient inference.}
	\label{fig:concept}
\end{figure}

To reconcile the efficiency of regression with the geometric fidelity of generative models, we propose the Physically-Grounded Manifold Projection (PGMP) framework. Distinct from standard diffusion, we reframe the MAR task by strategically adapting the Direct $x$-Prediction paradigm. Drawing theoretical grounding from recent advances in Consistency Models \cite{song2023consistency} and Rectified Flows \cite{liu2022flow}, we posit that the clean anatomical image lies on a low-dimensional manifold $\mathcal{M}$ embedded in the ambient artifact space. Instead of iteratively predicting the noise or velocity field, our model learns to project the corrupted observation directly onto this anatomical manifold in a single deterministic step. This approach effectively bridges the gap between regression efficiency and generative quality: it retains the speed of deterministic networks while leveraging the manifold constraints inherent in generative modeling to preserve high-frequency details.

However, applying this generative paradigm to dental imaging requires addressing specific medical constraints regarding data fidelity and semantic validity. To this end, our framework integrates three synergistic innovations. First, addressing the domain gap at its source, we establish an Anatomically-Adaptive Physics Simulation (AAPS) pipeline. Unlike simplistic monochromatic methods, AAPS incorporates polychromatic spectral modeling and a volumetric placement strategy for metal implants, synthesizing artifacts that faithfully replicate the non-linear beam hardening observed in real clinical scanners. Second, to execute the projection, we design the DMP-Former, an isotropic Vision Transformer backbone equipped with Rotary Positional Embeddings (RoPE). This architecture preserves the full-spectrum spatial resolution essential for resolving trabecular micro-structures, avoiding the information loss typical of hierarchical downsampling. Finally, to curb the risk of hallucination, we introduce a Semantic-Structural Alignment (SSA) mechanism. Leveraging MedDINOv3 \cite{li2025meddinov3}—a domain-adapted medical foundation model—as a digital anatomist, we distill expert-level priors into the restoration network, enforcing that the projected anatomy aligns with distinct biological semantics.

Our main contributions are summarized as follows:
\begin{itemize}
    \item We establish an Anatomically-Adaptive Physics Simulation (AAPS) pipeline that models complex polychromatic interactions to create high-fidelity digital twins, thereby minimizing the domain shift between synthetic training and clinical application.
    \item We adapt the Direct Manifold Projection paradigm to dental MAR via the DMP-Former, which overcomes the inefficiency of stochastic sampling to achieve high-fidelity restoration with superior inference speed.
    \item We devise a Semantic-Structural Alignment (SSA) mechanism that incorporates medical foundation priors (MedDINOv3) to effectively constrain the solution space, ensuring the rigorous preservation of diagnostic-critical details.
    \item We conduct comprehensive validation via downstream segmentation tasks and expert-aligned clinical assessment, demonstrating the framework's superior generalizability and clinical utility compared to state-of-the-art methods.
\end{itemize}

%% ===============================================
%% Section 2: Related Work
%% ===============================================

%% ===============================================
%% Section 2: Related Work (Merged Version)
%% ===============================================
\section{Related Work}
\label{sec:related_work}

Research on Metal Artifact Reduction (MAR) has evolved to balance physical fidelity with perceptual quality. We review this landscape by first analyzing the trade-offs between supervised regression and stochastic diffusion, then introducing the deterministic generative paradigms that ground our approach, and finally discussing the role of medical foundation models in semantic guidance.

\subsection{Evolution of Reconstruction: From Regression to Stochastic Diffusion}
The MAR field has historically been dominated by Supervised Learning, progressing from image-domain CNNs (CNNMAR \cite{zhang2018convolutional}) to physics-informed dual-domain networks (DuDoNet \cite{lin2019dudonet}, InDuDoNet+ \cite{wang2023indudonet}). While efficient, these methods primarily minimize pixel-wise objectives (L1/L2). As noted in recent benchmarks \cite{wang2023semimar}, this leads to the ``regression-to-the-mean'' phenomenon, resulting in waxy, over-smoothed textures that obliterate fine trabecular details. Furthermore, their reliance on paired data creates a substantial ``synthetic-to-real'' domain gap when training on simplistic simulations.

To address texture loss, Denoising Diffusion Probabilistic Models (DDPMs) have emerged as the new state-of-the-art \cite{karageorgos2024denoising, zhang2025coupled}. Methods like DuDoDp \cite{liu2024unsupervised} treat MAR as a conditional generation task, achieving superior perceptual realism. However, standard diffusion operates via an Iterative Noise Prediction ($\epsilon$-prediction) paradigm, requiring hundreds of stochastic sampling steps ($T \gg 100$). We argue this is theoretically suboptimal for medical restoration: it incurs prohibitive computational latency and introduces aleatoric uncertainty, where stochastic sampling variations can compromise the strict reproducibility required for diagnosis \cite{ma2025radiologist, bhutto2023denoising}.

\subsection{Deterministic Generative Priors via Flow Matching}
To reconcile the efficiency of regression with the quality of generation, recent theoretical advances have shifted focus toward Deterministic Generative Models. Frameworks such as Flow Matching \cite{lipman2022flow}, Rectified Flows \cite{liu2022flow}, and Consistency Models \cite{song2023consistency} demonstrate that the complex trajectory of diffusion can be distilled into a straight-line probability path, allowing for the Direct Prediction of Data ($x$) rather than noise. Our work strategically adapts this Direct $x$-Prediction paradigm to the inverse problem of MAR. Aligning with the Manifold Assumption explored in recent ``Just image Transformer'' (JiT) philosophies \cite{li2025back}, we formulate MAR as a single-step projection onto a learned anatomical manifold. Unlike standard regression which approximates a statistical average, this approach leverages the geometric constraints of generative models to recover high-frequency details without the computational burden or stochastic instability of iterative SDE solvers.

\subsection{Medical Foundation Models as Semantic Anchors}
While geometric restoration addresses texture, ensuring semantic plausibility remains a challenge, especially for unsupervised or limited-data scenarios where networks are prone to structural hallucinations \cite{zhang2025coupled}. Learning robust anatomical priors from scratch is inefficient. Recently, Vision Foundation Models have revolutionized feature representation. However, self-supervised pioneers like DINOv2 \cite{oquab2023dinov2}, trained on natural RGB images, suffer from a modality gap when applied to radiological densities (Hounsfield Units). To bridge this gap, domain-adapted encoders such as MedDINOv3 \cite{li2025meddinov3} and DINOv3 \cite{simeoni2025dinov3} have emerged, establishing expert-level feature priors. Incorporating insights from representation alignment theories (e.g., iREPA \cite{singh2025matters}), our work synergizes manifold projection with these medical foundation priors. By using MedDINOv3 as a digital anatomist, we constrain the restoration process to biologically valid anatomies, effectively curbing the risk of hallucination inherent in pure generative approaches.

%% Insertion of Figure 1-method (Validation Metrics)
\begin{figure}[t]
	\centering
	\includegraphics[width=1.0\linewidth]{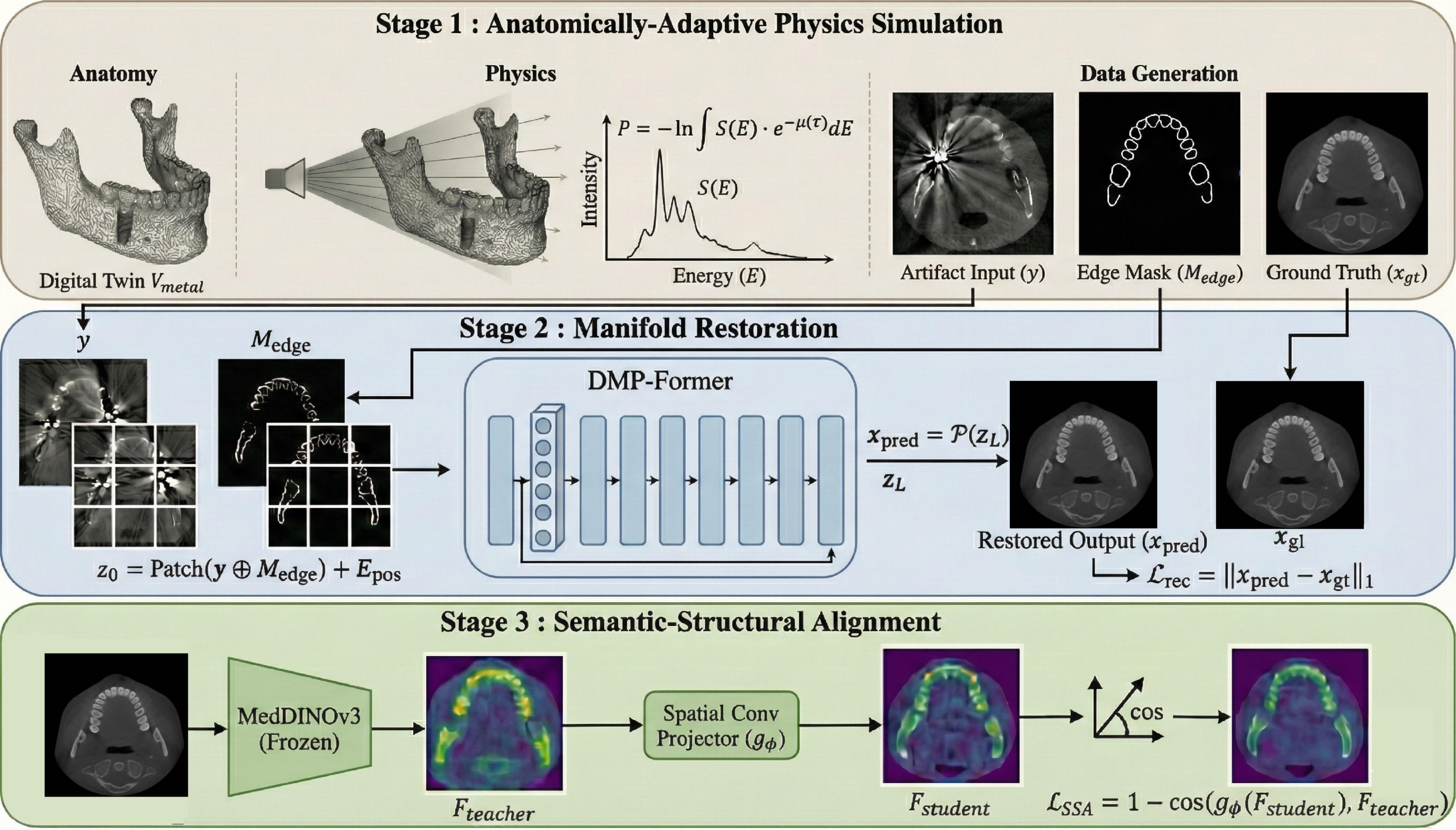}
    \caption{Overview of the Hierarchical PGMP Framework. The pipeline is structured into three phases:
    (Top) Anatomically-Adaptive Physics Simulation (AAPS):
    We generate realistic training pairs by simulating polychromatic X-ray attenuation on patient-specific digital twins ($V_{metal}$), bridging the domain gap between synthetic and clinical data.
    (Middle) DMP-Former Network:
    The student network utilizes an isotropic ViT backbone with structural conditioning (AdaLN-Zero) to directly project artifact-corrupted inputs ($y$) onto the clean anatomical manifold ($x_{pred}$), bypassing iterative denoising.
    (Bottom) Semantic-Structural Alignment (SSA):
    To ensure diagnostic fidelity, a frozen medical foundation model (MedDINOv3) acts as a teacher. It guides the student network to capture expert-level anatomical semantics by minimizing the feature divergence ($\mathcal{L}_{SSA}$) between the restored and ground-truth volumes.
    }
    \label{fig:framework_overview}
\end{figure}

%% ===============================================
%% Section 3: Methodology
%% ===============================================
\section{Methodology}
\label{sec:method}

We propose the Physically-Grounded Manifold Projection (PGMP) framework, a hierarchical approach designed to rigorously address the dual challenges of domain generalization and structural preservation in Metal Artifact Reduction. As visualized in Fig. \ref{fig:framework_overview}, the pipeline operates through three synergistic stages, transitioning from physical data synthesis to geometric restoration and finally to semantic refinement. First, to bridge the synthetic-to-real domain gap, we employ the Anatomically-Adaptive Physics Simulation (AAPS), which synthesizes high-fidelity training pairs via polychromatic spectral modeling. Second, building upon this robust data substrate, the Direct Manifold Projection Transformer (DMP-Former) executes a deterministic mapping from the corrupted observation to the clean anatomical manifold, ensuring inference efficiency. Finally, to guarantee clinical plausibility, the Semantic-Structural Alignment (SSA) mechanism leverages medical foundation priors to enforce expert-level consistency. In this section, we detail the theoretical formulation and implementation of each component.

\subsection{Physically-Driven Clinical Data Synthesis}
The primary bottleneck limiting current supervised MAR methods is the substantial distribution shift between training data and clinical acquisitions. Conventional simulations often rely on monochromatic ray-tracing and simplistic 2D masks, failing to capture the non-linear beam hardening and complex scattering effects inherent in real-world scanners. To mitigate this issue, we introduce the Anatomically-Adaptive Physics Simulation (AAPS) pipeline. Unlike naive approaches, AAPS constructs volumetrically consistent ``Digital Twins'' and simulates the stochastic physical processes of X-ray acquisition using spectral imaging theory. This ensures that the generated artifacts respect both the geometric constraints of dental restorations and the polychromatic characteristics of clinical CBCT systems.

\subsubsection{Construction of Anatomically-Adaptive Digital Twins}
The generation of realistic artifacts fundamentally depends on the accurate placement of metal prosthetics within the patient's anatomical context. Traditional methods, such as those used in early deep learning baselines \cite{zhang2018convolutional}, often randomly insert geometric primitives (e.g., cylinders) into isolated 2D slices. This approach ignores the inherent 3D continuity of dental structures, resulting in artifacts that are disjointed and geometrically impossible in a clinical setting, which exacerbates the domain gap \cite{wang2023semimar}.

To resolve this, we implement a stochastic yet clinically-grounded simulation planner to determine the morphology and location of metal objects. The planner analyzes 3D tooth segmentation masks—derived from high-quality annotations in the MICCAI STS benchmark \cite{wang2026miccai, wang2024semi}—to compute the vertical extent of each tooth based on FDI notation. Based on a probabilistic model of adult dental restoration prevalence, we assign a specific restoration state (e.g., Sound, Filled, Crowned, Implant, or Bridge) to each tooth. To ensure \textit{Volumetric Fidelity}, metal components are not placed arbitrarily but are confined to clinically valid zones relative to the tooth's height $h$. For instance, implants are restricted to the alveolar bone region ($0.0h - 0.60h$), whereas crowns occupy the occlusal third ($0.40h - 1.0h$), and fillings are distributed across the coronal and middle thirds ($0.40h - 0.95h$).

\begin{figure}[t]
	\centering
	\includegraphics[width=1.0\linewidth]{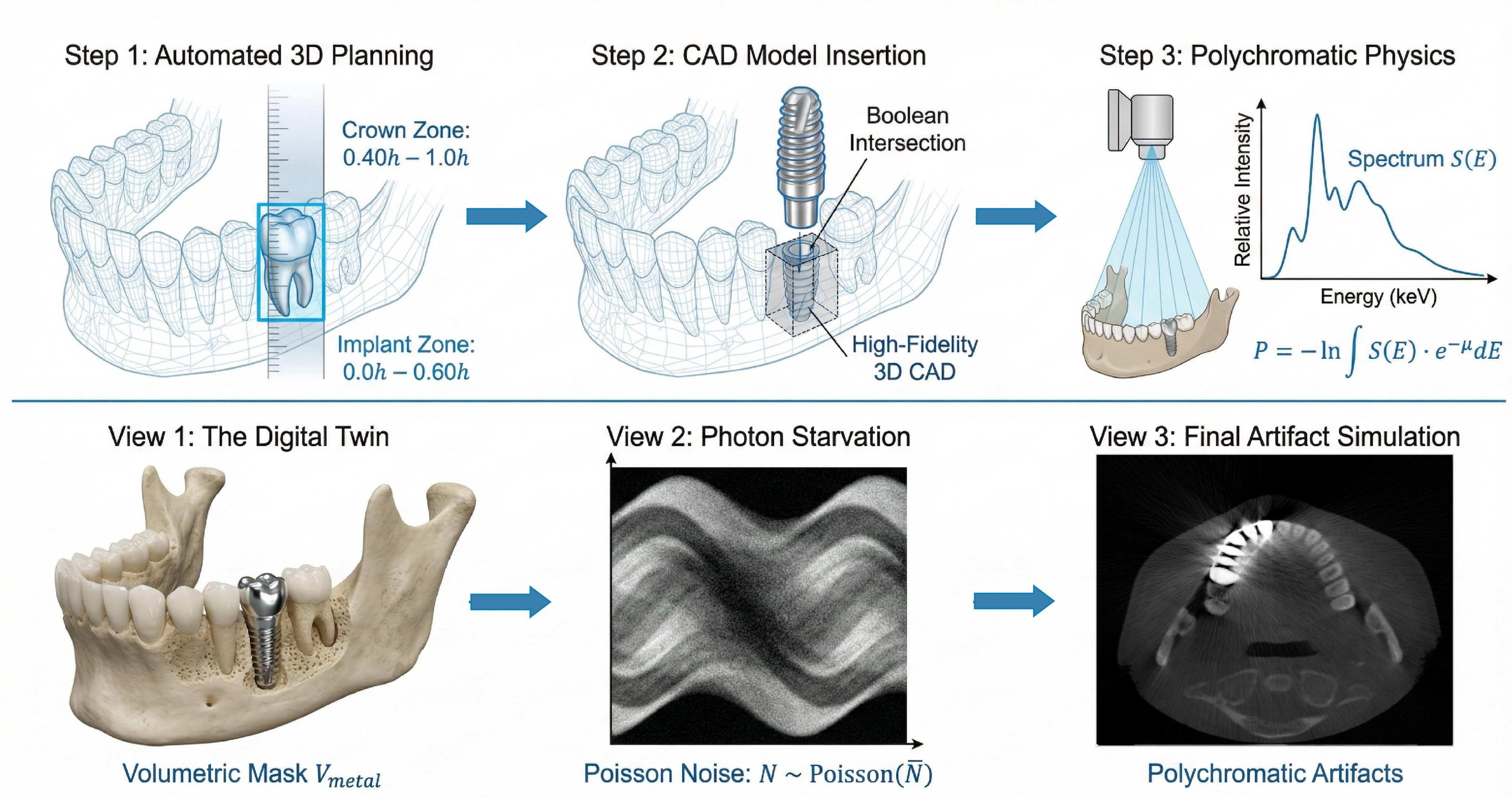}
	\caption{Workflow of the proposed Anatomically-Adaptive Physics Simulation (AAPS). Step 1 \& 2: Automated 3D planning inserts high-fidelity CAD models into clinically valid zones (e.g., Implant Zone $0.0h-0.60h$) to generate a volumetrically consistent metal mask $V_{metal}$. Step 3: The physics simulation incorporates a polychromatic X-ray spectrum $S(E)$ to model beam hardening according to the integral $P = -\ln \int S(E) \cdot e^{-\mu(E)L} dE$. Views: The process results in a digital twin that exhibits realistic photon starvation (Poisson noise $N \sim \text{Poisson}(\bar{N})$) and characteristic dark band artifacts.}
	\label{fig:aaps_pipeline}
\end{figure}

Instead of projecting static 2D shapes, we utilize a library of high-fidelity 3D CAD models. For each planned restoration, the standard mesh is rigidly aligned and anisotropically scaled to fit the patient's specific tooth morphology. We then perform a dynamic boolean intersection between the 3D mesh and the axial plane. This process guarantees that the resulting metal mask $V_{metal}$ exhibits continuous, anatomically-compliant changes along the Z-axis—such as the gradual transition from a screw implant's circular apex to its threaded body. This volumetric consistency is physically critical, as it allows the subsequent simulation to generate realistic, spatially-propagating streak artifacts that accurately reflect the complex scattering geometry observed in clinical scans \cite{zhang2025coupled}. The overall pipeline is illustrated in Fig. \ref{fig:aaps_pipeline}.

% Insertion of Figure 2

\subsubsection{Polychromatic Physics Modeling with Scattering Approximation}
The second phase of AAPS bridges the synthetic-to-real domain gap by systematically emulating the complex stochastic and deterministic physical processes inherent in clinical CBCT acquisition. In stark contrast to conventional MAR simulation pipelines that rely on a monochromatic Beer–Lambert law with a fixed energy $E_0$ \cite{lin2019dudonet, zhang2018convolutional}, real-world CBCT systems employ polychromatic X-ray spectra . This polychromaticity induces non-linear attenuation phenomena that simple simulations fail to capture. To approximate these effects with high fidelity, we implement a multi-stage projection pipeline grounded in spectral imaging theory \cite{bushberg2011essential}.

We model the X-ray source as an energy-dependent spectrum $S(E)$, calibrated to a standard clinical 120 kVp tube voltage with 2 mm Aluminum filtration. The projection intensity $P$ is computed by integrating over the entire energy spectrum, explicitly accounting for the energy-dependent attenuation coefficients $\mu_m(E)$ of each constituent material $m \in \{\text{water, bone, metal}\}$:
\begin{equation}
P = -\ln \left( \frac{1}{I_0} \int_{E} S(E) \cdot \exp\left(-\sum_{m} \int_{L} x_m(r) \cdot \mu_m(E) \, dr\right) \, dE \right)
\end{equation}
where $x_m(r)$ denotes the spatial density map of material $m$ derived via threshold-based decomposition of Hounsfield units, and $L$ represents the ray path.  This formulation explicitly captures the \textit{spectral hardening} effect: since low-energy photons are preferentially absorbed by high-Z metal implants, the effective beam energy increases along the path \cite{klotz1987reduction}. This process naturally generates the characteristic dark bands and cupping artifacts observed in clinical scans, which are intrinsically absent in monochromatic simulations.

Following the deterministic attenuation, we model the stochastic loss of photons. Behind dense metallic objects, the transmitted photon flux drops dramatically, leading to \textit{photon starvation}—a regime where quantum noise dominates the signal. To simulate this, we first compute the expected photon count at each detector bin as $\bar{N} = N_0 \cdot \exp(-P)$, with $N_0 = 5 \times 10^5$ photons/bin representing a clinically realistic flux. The actual detected photon count $N$ is then sampled from a Poisson distribution:
\begin{equation}
N \sim \text{Poisson}(\bar{N})
\end{equation}
This is followed by a thresholding step $N \leftarrow \max(N, N_{min})$ (where $N_{min}=10$) to prevent log-domain overflow while preserving the statistical signature of extreme low-signal regions.

While full Monte Carlo (MC) particle transport represents the gold standard for simulating X-ray scattering, its immense computational cost renders it prohibitive for generating large-scale training datasets \cite{zhang2025coupled}. To reconcile physical fidelity with computational feasibility, we adopt a convolutional scattering proxy that approximates the spatial distribution of scattered photons. Based on the observation that scatter primarily adds a low-frequency bias to the projection, we estimate the scatter fluence $I_{scatter}$ as:
\begin{equation}
I_{scatter} = SPR \cdot \sum_{m} (I_{primary, m} * \kappa_\sigma)
\end{equation}
where $*$ denotes 2D convolution, and $\kappa_\sigma$ is a radially symmetric Gaussian kernel with standard deviation $\sigma=10$ pixels. The Scatter-to-Primary Ratio ($SPR$) is set to 0.1 based on empirical clinical values. Although this proxy simplifies the complex angular dependencies of Compton scattering, it effectively captures the ``blooming'' effects and low-frequency haze that are critical for training robustness, without the prohibitive cost of ray-by-ray particle tracking.

Finally, we inject electronic readout noise modeled as additive Gaussian noise with standard deviation $\sigma_e$, yielding the final noisy sinogram $N_{final} = N + \mathcal{N}(0, \sigma_e^2)$. The sinogram is then reconstructed using a filtered backprojection (FBP) algorithm. 
% To validate the fidelity of our simulation, we employed MedDINOv3 \cite{li2025meddinov3} to compare the feature space distribution of AAPS-generated data with that of real clinical CBCT scans, confirming that our physics-driven approach successfully captures the high-order statistical characteristics of clinical artifacts necessary for effective domain adaptation.

\subsection{Manifold-based Artifact Reduction Network}
While diffusion models have revolutionized generative imaging through iterative refinement \cite{ho2020denoising, song2020score}, their standard stochastic formulation presents a theoretical misalignment with the requirements of medical restoration. Specifically, the iterative removal of noise is computationally prohibitive, and the stochastic sampling trajectory introduces aleatoric uncertainty incompatible with diagnostic reproducibility. In this section, we reframe the Metal Artifact Reduction (MAR) task by leveraging the theoretical framework of Flow Matching \cite{lipman2022flow, albergo2023stochastic}. By strategically adapting the Direct $x$-Prediction parameterization—previously validated in generic generative contexts like Consistency Models \cite{song2023consistency}—we construct a restoration network that combines the geometric fidelity of generative models with the deterministic efficiency required for clinical deployment.

\subsubsection{Formulation: Deterministic Projection via Rectified Flows}
We formulate the MAR process as traversing a probability path that connects the distribution of artifact-corrupted inputs to the distribution of clean anatomical volumes. Unlike standard generative models where the source distribution is pure Gaussian noise, here the source is the structured artifact domain. Drawing upon the theory of Rectified Flows \cite{liu2022flow}, we define a linear interpolation schedule $z_t$ at time $t \in [0, 1]$:
\begin{equation}
z_t = t x + (1 - t) y
\end{equation}
where $z_1 = x$ represents the clean anatomical data lying on a low-dimensional manifold $\mathcal{M} \subset \mathbb{R}^N$, and $z_0 = y$ represents the corrupted observation in the high-dimensional ambient space. Note that in this context, the corruption term $(y - x)$ corresponds to the structured metal artifacts rather than isotropic Gaussian noise. The trajectory of this restoration process is governed by a flow velocity field $v_t$, defined as the time derivative of the state:
\begin{equation}
v_t = \frac{d z_t}{d t} = x - y
\end{equation}
Standard approaches, such as Conditional Flow Matching (CFM) \cite{lipman2022flow}, typically train a network to approximate this velocity field $v_t$ and then solve an Ordinary Differential Equation (ODE) over many steps ($T \gg 1$) to generate samples. However, for the inverse problem of MAR, the goal is not to sample diversely from a noise distribution, but to recover the \textit{unique, deterministic} anatomical truth $x$ given the observation $y$. The discretization errors inherent in multi-step ODE solvers can lead to structural drift \cite{karras2022elucidating}. Therefore, we bypass the intermediate velocity estimation and explicitly model the restoration as a single-step projection onto the manifold $\mathcal{M}$.

\subsubsection{The Direct $x$-Prediction Paradigm}
Leveraging the linear relationship in Eq. (4), we adopt the Direct $x$-Prediction Paradigm, where the network $f_\theta$ is trained to predict the clean volume $x$ directly from the corrupted input $y$. While mathematically analogous to single-step regression, we theoretically frame this as a Manifold Projection task rather than simple image-to-image translation.

This distinction is grounded in the Manifold Assumption \cite{bengio2013representation, li2025back}. In traditional regression (e.g., U-Nets), the network often learns a statistical average of plausible solutions, leading to the ``waxy'' textures criticized in prior work \cite{wang2023semimar}. In our framework, we posit that valid anatomical images reside on a lower-dimensional manifold $\mathcal{M}$. By formulating the objective as a direct mapping $f_\theta: Y \rightarrow \mathcal{M}$, and reinforcing it with structural conditioning $M_{edge}$, we constrain the output space to biologically valid structures. This allows the network to focus its capacity on reconstructing high-frequency deterministic semantics—such as enamel boundaries and trabecular patterns—rather than modeling the high-entropy ambient artifact space \cite{salimans2022progressive}. Consequently, the DMP-Former is optimized via a direct manifold reconstruction objective:
\begin{equation}
\mathcal{L}_{manifold} = \| f_\theta(y, M_{edge}) - x_{gt} \|_1
\end{equation}
where $M_{edge}$ provides the necessary structural priors to guide the projection, ensuring the solution remains within the boundaries of dental anatomy.

% Insertion of Figure 4
\begin{figure}[t]
	\centering
	\includegraphics[width=1.0\linewidth]{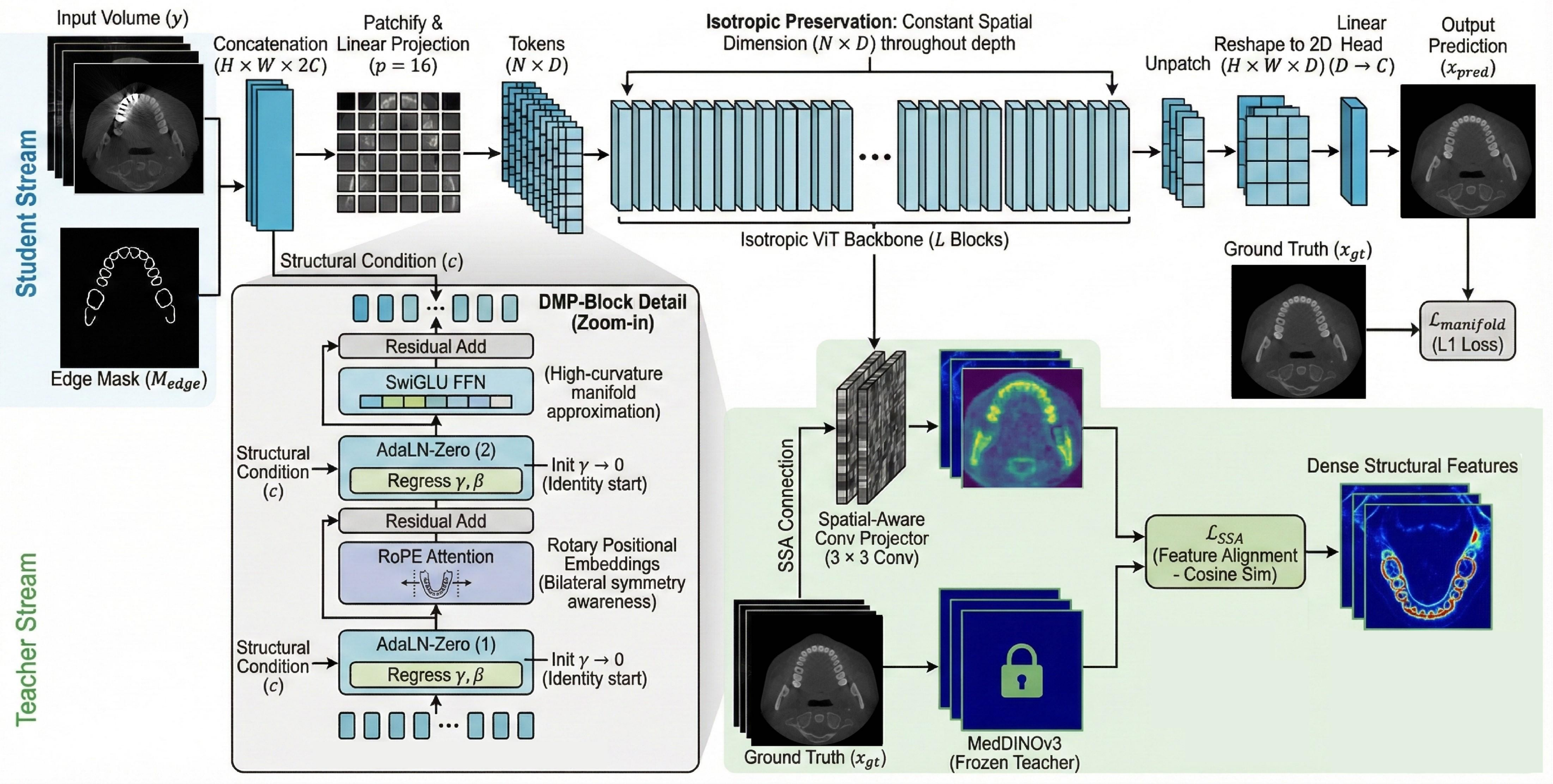}
	\caption{Schematic overview of the Direct Manifold Projection Transformer (DMP-Former). The network processes the artifact-corrupted volume $y$ concatenated with the structural edge mask $M_{edge}$. Unlike hierarchical U-Nets, it employs an isotropic ViT backbone with $L$ stacked DMP-Blocks. The zoom-in detail illustrates the integration of \textit{AdaLN-Zero} for condition injection (regressing scale $\gamma$ and shift $\beta$), \textit{RoPE Attention} for relative spatial awareness, and \textit{SwiGLU} for expressive feed-forward dynamics. The model directly predicts the clean anatomy $x_{pred}$ via the $x$-prediction paradigm, minimizing the manifold reconstruction loss $\mathcal{L}_{manifold}$, while intermediate features connect to the SSA mechanism for semantic guidance.}
	\label{fig:dmp_former}
\end{figure}

\subsubsection{DMP-Former: Isotropic Architecture with Structural Conditioning}
To execute the manifold projection with high geometric fidelity, we architect the DMP-Former based on the ``Just image Transformer'' (JiT) philosophy \cite{li2025back}. As visualized in Fig. \ref{fig:dmp_former}, distinct from hierarchical U-Net architectures commonly used in medical imaging \cite{ronneberger2015u, wang2023indudonet} which progressively downsample spatial resolution to expand the receptive field, we employ an isotropic Vision Transformer (ViT) backbone \cite{dosovitskiy2020image}. This design choice is strictly motivated by the clinical necessity of dental imaging: key diagnostic features, such as the periodontal ligament space ($<0.2$ mm) and trabecular bone micro-architecture, represent high-frequency details that are easily obliterated by the pooling operations inherent in hierarchical models. By maintaining a constant high-resolution feature map throughout all $L$ layers, our isotropic design ensures that these subtle anatomical signals are preserved losslessly from input to output, addressing the "over-smoothing" limitation of traditional regression networks.

The restoration process initiates by forming a composite input tensor, where the artifact-corrupted volume $y$ is concatenated channel-wise with the structural tooth edge mask $M_{edge}$. This tensor is tokenized into non-overlapping patches of size $p \times p$ via a convolutional projection, embedding the physical priors directly into the initial latent space. These tokens are then processed through a stack of DMP-Blocks. To mathematically formalize the information flow, let $h_{l-1}$ denote the input tokens to the $l$-th block and $c$ represent the condition embedding derived from the structural mask. The forward pass of each block is governed by the following transition equations:
\begin{align}
\hat{h}_l & = \text{Attn}(\text{AdaLN}(h_{l-1}, c)) + h_{l-1} \\
h_l & = \text{FFN}(\text{AdaLN}(\hat{h}_l, c)) + \hat{h}_l
\end{align}
This formulation integrates three specialized mechanisms designed to tailor the standard transformer for the ill-posed MAR task. First, to strictly enforce the structural guidance provided by $M_{edge}$, we replace standard normalization with Adaptive Layer Normalization (AdaLN-Zero), a technique adapted from scalable diffusion transformers \cite{peebles2023scalable}. As shown in the equations, the function $\text{AdaLN}(h, c) = \gamma_c \cdot \text{Norm}(h) + \beta_c$ dynamically modulates the feature statistics of the normalized tokens using scale $\gamma_c$ and shift $\beta_c$ parameters, which are regressed directly from the condition $c$. Crucially, we initialize the regression weights such that $\gamma_c$ starts at zero. This initialization causes the entire block to behave as an identity function at the beginning of training, allowing the network to first learn a coarse identity mapping before progressively injecting complex structural modifications. This strategy significantly stabilizes the optimization of the deep projection mapping, preventing the gradient explosions often seen in direct inverse problem solving.

Simultaneously, to address the spatial ambiguity inherent in patch-based processing, we integrate Rotary Positional Embeddings (RoPE) \cite{su2024roformer} within the self-attention mechanism ($\text{Attn}$). Unlike absolute position encodings that fix tokens to specific grid coordinates, RoPE encodes spatial information by rotating the query and key vectors in the embedding space. This relative positioning is particularly advantageous for dental anatomy, which exhibits strong bilateral symmetry (e.g., the left molar structurally mirrors the right molar). RoPE enables the attention heads to explicitly capture these relative geometric distances, allowing the model to utilize information from the contralateral healthy tooth to infer the structure of the occluded region. Finally, the feature transformation within the Feed-Forward Network ($\text{FFN}$) is powered by the SwiGLU activation function \cite{shazeer2020glu}. By introducing a gated mechanism, SwiGLU offers a significantly more expressive non-linear variation compared to the standard GELU. This enhanced non-linearity is essential for approximating the highly complex, high-curvature manifold of dental data, enabling the network to disentangle the non-linear beam-hardening artifacts from the underlying biological tissue with greater precision. The processed tokens are ultimately unpatched and projected linearly to pixel space to yield the clean volume $x_{pred}$, while intermediate features are tapped for semantic alignment.

\subsection{Semantic-Structural Alignment by Medical Foundation Priors}
While the DMP-Former achieves considerable inference efficiency through direct manifold projection, relying solely on pixel-space reconstruction objectives (e.g., L1 loss) entails a significant theoretical risk. As established in inverse problem theory, minimizing pixel-wise error in ill-posed tasks often leads to ``regression to the mean,'' resulting in outputs that are statistically safe but perceptually blurry \cite{wang2023semimar}. In medical imaging, this ambiguity is dangerous: a network might effectively minimize L1 error while failing to respect intrinsic biological constraints, such as the continuity of enamel boundaries or the trabecular orientation. This disconnect can lead to structural hallucinations, where the restored image is visually smooth but anatomically implausible \cite{bhutto2023denoising}. To mitigate this, we introduce the Semantic-Structural Alignment (SSA) mechanism. Acting as a deterministic substitute for the discriminator in GANs, SSA leverages a medical foundation model as a ``Digital Anatomist,'' guiding the student network’s feature space toward clinically valid representations without introducing adversarial instability.

\subsubsection{Distilling Domain-Specific Anatomical Semantics}
A critical prerequisite for effective alignment is the choice of the teacher model. Prior generative approaches have often relied on generic vision encoders, such as DINOv2 \cite{oquab2023dinov2}, which are pretrained on natural images (ImageNet). We argue that such models suffer from a fundamental modality mismatch: they are optimized to discriminate macroscopic surface properties of objects (e.g., texture, color) under natural illumination. In contrast, radiological diagnosis relies on internal tissue density (Hounsfield Units) and subtle morphological variations. Using a generic encoder can therefore lead to feature misalignment.

To overcome this, we employ MedDINOv3 \cite{li2025meddinov3}, a state-of-the-art foundation model adapted from DINOv3 \cite{simeoni2025dinov3} and pretrained on over 3 million clinical CT slices. Unlike RGB-based models, MedDINOv3 encodes rich, domain-specific priors, enabling it to robustly discriminate between tissues with overlapping intensities—for instance, distinguishing true cortical bone from the high-density streak artifacts that confuse standard networks. We treat MedDINOv3 as a frozen teacher network $\Phi_{teach}$ to extract dense semantic features from the clean ground truth target $x_{gt}$:
\begin{equation}
F_{teach} = \Phi_{teach}(\text{Norm}(x_{gt}))
\end{equation}
where $\text{Norm}(\cdot)$ denotes the standard windowing and normalization. By aligning the student's latent features with $F_{teach}$, the DMP-Former is compelled to learn representations that are linearly separable in the medical semantic space, ensuring that the restored content adheres to radiological logic rather than merely satisfying statistical regularities.

\subsubsection{Spatial-Aware Alignment via iREPA Principles}
Standard representation alignment, or REPA \cite{yu2024representation}, typically utilizes global pooling to match feature distributions. However, such operations inadvertently obfuscate the fine-grained spatial details essential for dental diagnostics—a global descriptor cannot capture the integrity of a 0.2 mm periodontal ligament. Drawing inspiration from the iREPA strategy \cite{singh2025matters}, which posits that preserving spatial topology (pairwise patch similarity) is critical for generation quality, we redesign the alignment pipeline with two spatially-aware components, as integrated into the workflow shown in Fig. \ref{fig:dmp_former}.

First, distinct from the spatially-agnostic MLPs used in standard distillation, we introduce a lightweight $3 \times 3$ convolutional projector $\mathcal{P}_{conv}$ to map the student's features (extracted from the deep layers of DMP-Former, $F_{student}$) to the teacher's feature dimension. This design strictly preserves the 2D grid topology, ensuring that alignment occurs at the pixel-aligned patch level rather than the image level:
\begin{equation}
F'_{student} = \mathcal{P}_{conv}(F_{student})
\end{equation}
Second, to prevent the alignment loss from being dominated by the extreme intensity outliers characteristic of metal artifacts, we implement a robust spatial normalization. We subtract the mean and scale by the variance computed across the spatial dimensions $(H, W)$ for each channel:
\begin{equation}
\hat{F} = \frac{F - \mu_{spatial}(F)}{\sigma_{spatial}(F) + \epsilon}
\end{equation}
This normalization suppresses global intensity biases—conceptually similar to Instance Normalization \cite{ulyanov2016instance}—and forces the optimization to focus on relative structural variations (morphology) rather than absolute pixel intensities. This is particularly crucial for MAR, as it encourages the restoration of tooth boundaries independent of the global brightness shifts caused by spectral hardening, ensuring that the alignment loss drives geometric recovery rather than simple brightness matching.

%% ===============================================
%% Section 3.4: Optimization
%% ===============================================
\subsection{Holistic Optimization via Multi-Scale Constraints}
\label{sec:loss}

To transform the ill-posed inverse problem of Metal Artifact Reduction into a tractable learning task, we construct a composite objective function that transcends simple pixel-wise regression. We argue that a clinically robust restoration model must satisfy constraints across three distinct abstraction levels: low-level signal fidelity, mid-level structural topology, and high-level semantic consistency. Consequently, we design a holistic optimization landscape where the DMP-Former is simultaneously guided by manifold reconstruction, semantic alignment, and gradient consistency.

The semantic constraint is enforced via the Semantic-Structural Alignment (SSA) loss. As detailed in the previous section, mere Euclidean distance in pixel space fails to capture the biological essence of anatomical structures. To rectify this, we rigorously define $\mathcal{L}_{SSA}$ as the negative cosine similarity  between the spatially normalized feature maps of the student and the teacher networks. By averaging this similarity across all spatial locations, we penalize feature divergence in the latent manifold:
\begin{equation}
	\mathcal{L}_{SSA} = 1 - \frac{\langle \hat{F}'_{student}, \hat{F}_{teach} \rangle}{\| \hat{F}'_{student} \|_2 \cdot \| \hat{F}_{teach} \|_2}
\end{equation}
This formulation acts as a semantic regularizer, compelling the network to prioritize the recovery of discriminative features—such as the texture of trabecular bone—over the blind removal of high-intensity pixels. Unlike adversarial losses in GANs which can induce mode collapse, $\mathcal{L}_{SSA}$ provides a stable, deterministic gradient toward the manifold of valid dental anatomy defined by the teacher model.

Complementary to the semantic prior, we introduce a geometric constraint to preserve the high-frequency diagnostic harmonics often eroded by L1 minimization. We incorporate an auxiliary Edge Consistency Loss ($\mathcal{L}_{edge}$), calculated as the L1 difference between the Sobel gradients  of the prediction and the ground truth. Crucially, this loss is spatially gated using the structural mask $M_{edge}$ derived from our AAPS simulation. By restricting the gradient penalty to the tooth Region of Interest (ROI), we focus the model's capacity on sharpening the distinct boundaries of the root canal and periodontal ligament, rather than wasting optimization effort on background noise.

The final optimization objective is a weighted synergy of these three complementary forces:
\begin{equation}
	\mathcal{L}_{total} = \underbrace{\| \hat{x} - x_{gt} \|_1}_{\mathcal{L}_{manifold}} + \lambda_{SSA} \underbrace{\mathcal{L}_{SSA}}_{\text{Semantic Prior}} + \lambda_{edge} \underbrace{\| \nabla \hat{x} - \nabla x_{gt} \|_{1, ROI}}_{\text{Structural Detail}}
\end{equation}
where the first term ensures basic signal fidelity on the clean manifold. Empirically, we balance these contributions by setting $\lambda_{SSA}=0.2$ and $\lambda_{edge}=0.1$. This multi-objective framework effectively constrains the solution space, ensuring that the final restoration achieves a clinically optimal equilibrium: visually clean (via $\mathcal{L}_{manifold}$), structurally sharp (via $\mathcal{L}_{edge}$), and biologically meaningful (via $\mathcal{L}_{SSA}$).

%% ===============================================
%% Section 4: Experiments
%% ===============================================
\section{Experiments}
\label{sec:exp}

\subsection{Experimental Setup}
\subsubsection{Datasets and Evaluation Protocols}
To rigorously evaluate the efficacy and robustness of the proposed PGMP framework, we established a dual-track evaluation protocol. This comprises a controlled, physically grounded synthetic benchmark for quantitative analysis and a diverse multi-center clinical dataset for assessing sim-to-real generalizability.

\paragraph{Physically-Grounded Training Benchmark (AAPS)}
We constructed the AAPS dataset utilizing the clean anatomical volumes from the STS24 dataset \cite{wang2026miccai}. Chosen for its high-resolution rendering of dental structures, STS24 provides the pristine ground truth essential for supervised learning. Leveraging the automated 3D implant planning algorithm described in Section \ref{sec:method}, we generated a comprehensive corpus of 9,441 paired training slices.

Crucially, to prevent data leakage and ensure rigorous evaluation, we implemented a strict patient-level split. Scans from a subset of distinct patients were reserved exclusively for testing ($N=1,153$ slices), ensuring that the network never encounters the anatomical identities used during training ($N=8,288$ slices). This separation guarantees that our downstream segmentation metrics reflect true generalization to unseen anatomy rather than overfitting to specific patient geometries. Furthermore, to prevent the network from exploiting simplistic geometric biases, we explicitly stratified the dataset to encompass the full spectrum of artifact morphologies defined by FDI dental notation. As detailed in Table \ref{tab:metal_stats}, the dataset includes a balanced distribution of 1,086 slices containing implants, 2,475 slices with crowns, and 5,880 slices with fillings, serving as the unified training benchmark for all supervised baselines to ensure fair comparison.

\begin{table}[htbp]
\centering
\caption{Detailed statistical analysis of geometric properties for metal artifacts across different restoration types. $P$-values derived from Kruskal-Wallis test. Post-hoc analysis performed using Mann-Whitney U test with Bonferroni correction, confirming statistically significant morphological differences between groups.}
\label{tab:metal_stats}
\resizebox{\columnwidth}{!}{
\begin{tabular}{lcccc}
\toprule
\textbf{Characteristic} & \textbf{Filled} & \textbf{Crowned} & \textbf{Implant} & \textbf{$P$-value\textsuperscript{a}} \\
\midrule
\multicolumn{5}{l}{\textit{Area (px$^2$)}} \\
\hspace{2mm} Mean $\pm$ SD & 579.70 $\pm$ 598.17 & 568.42 $\pm$ 785.96 & 757.05 $\pm$ 858.15 & \multirow{3}{*}{<0.001} \\
\hspace{2mm} Median [Q1, Q3] & 400 [255, 676] & 300 [70, 775] & 528 [241.75, 1023.75] & \\
\hspace{2mm} Range (Min--Max) & 24--6716 & 0--5041 & 1--6460 & \\
\addlinespace
\multicolumn{5}{l}{\textit{Width (px)}} \\
\hspace{2mm} Mean $\pm$ SD & 23.26 $\pm$ 10.22 & 19.17 $\pm$ 15.49 & 25.02 $\pm$ 13.53 & \multirow{3}{*}{<0.001} \\
\hspace{2mm} Median [Q1, Q3] & 21 [16, 28] & 17 [8, 30] & 24 [15, 33] & \\
\hspace{2mm} Range (Min--Max) & 5--73 & 0--71 & 1--85 & \\
\addlinespace
\multicolumn{5}{l}{\textit{Height (px)}} \\
\hspace{2mm} Mean $\pm$ SD & 21.37 $\pm$ 9.50 & 18.60 $\pm$ 14.73 & 24.14 $\pm$ 12.61 & \multirow{3}{*}{<0.001} \\
\hspace{2mm} Median [Q1, Q3] & 20 [15, 25] & 19 [8, 26] & 23 [15, 31] & \\
\hspace{2mm} Range (Min--Max) & 4--92 & 0--71 & 1--76 & \\
\addlinespace
\bottomrule
\end{tabular}
}
\end{table}

\paragraph{Real-World Clinical Benchmark}
Beyond synthetic validation, we established a rigorous benchmark for Cross-Protocol Generalizability using real clinical CBCT scans. This dataset was retrospectively collected from multiple clinical centers, explicitly selected to challenge the model's robustness against distribution shifts. It encompasses a wide variety of scanner manufacturers (e.g., Kavo, Planmeca) and acquisition protocols (varying FOV, kVp settings, and voxel resolutions) not seen during training. The collection focuses on complex cases with native metal artifacts where no ground-truth clean images exist. This explicit separation of training (synthetic) and testing (clinical) domains allows us to rigorously assess the ``Sim-to-Real'' transferability of our model and facilitate qualitative assessment by expert radiologists in real-world diagnostic scenarios.

%% Insertion of Figure 5 (Slice Analysis)
\begin{figure}[t]
	\centering
	\includegraphics[width=1.0\linewidth]{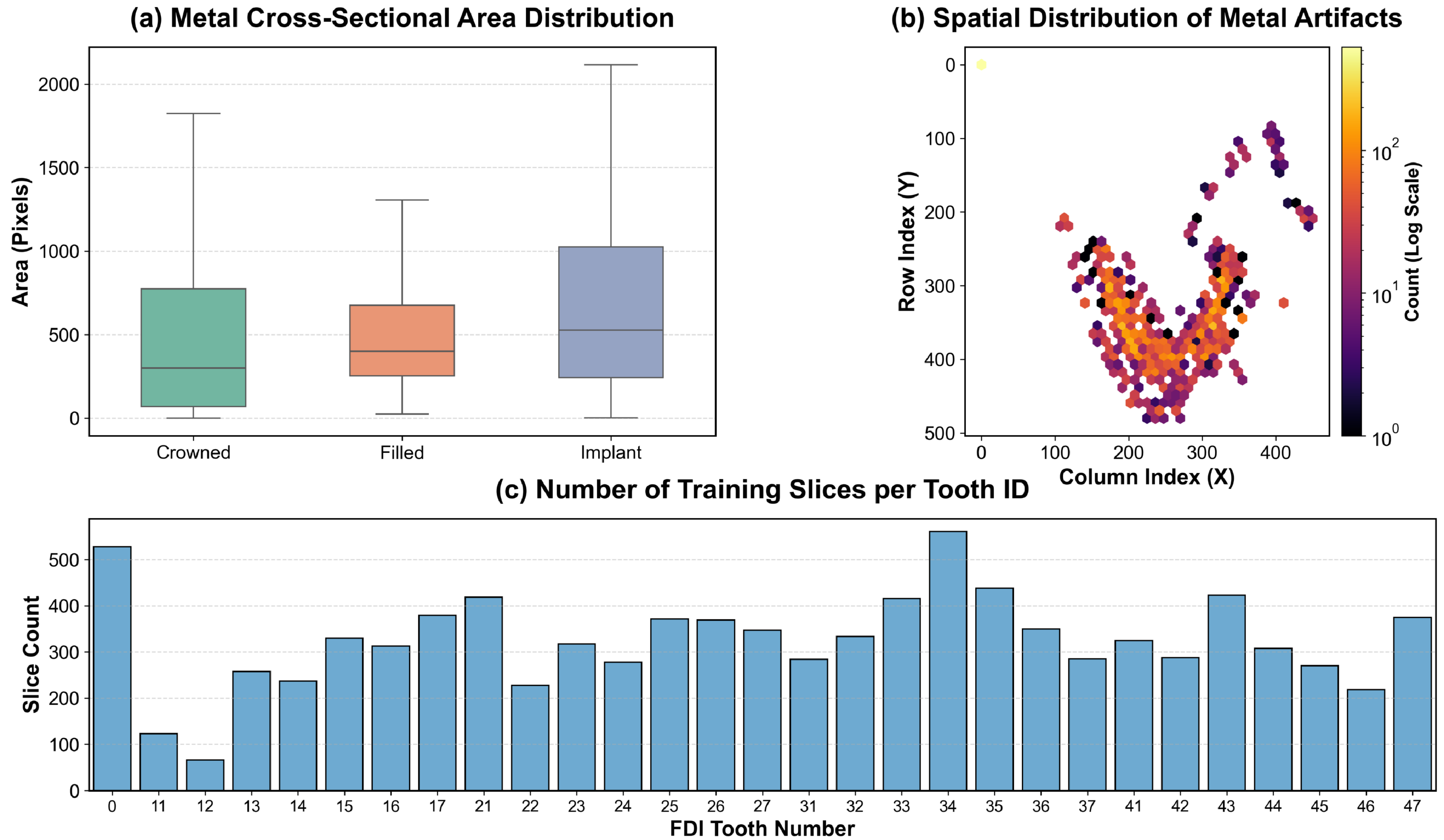}
	\caption{Statistical distribution analysis of the AAPS training dataset ($N=9441$). (a) Boxplot of metal cross-sectional areas across three restoration types. Implants exhibit a significantly larger area and variance ($p < 0.001$), presenting a harder restoration challenge. (b) Spatial heatmap of metal artifact centroids on the axial plane, demonstrating dense coverage of the dental arch curve without spatial bias. (c) Histogram of training slices per FDI tooth ID, confirming that the dataset includes diverse examples from both anterior (e.g., Incisors \#11, \#21) and posterior (e.g., Molars \#16, \#26) regions.}
	\label{fig:slice_analysis}
\end{figure}
\subsubsection{Implementation Protocols and Baseline Fairness}
The proposed PGMP framework was implemented in PyTorch and trained on a computing cluster equipped with NVIDIA RTX 4090 GPUs. To accommodate the memory demands of high-dimensional manifold projection, the DMP-Former processes non-overlapping patches of size $p=16$. We adopted the AdamW optimizer with momentum parameters $\beta_1 = 0.9$ and $\beta_2 = 0.999$ to stabilize the optimization trajectory within the transformer's complex loss landscape. The learning rate was initialized at $1 \times 10^{-4}$ and managed via a cosine annealing scheduler, ensuring smooth convergence towards the global minimum while preventing the model from becoming trapped in the sharp local optima associated with non-convex artifact patterns. The model was trained for 200 epochs with a batch size of 16. Regarding the hyperparameter balancing in the total objective function (Eq. 12), we empirically set the weights for the Semantic-Structural Alignment (SSA) loss and Edge Consistency loss to $\lambda_{SSA} = 0.2$ and $\lambda_{edge} = 0.1$, respectively, based on validation performance on a hold-out set. Notably, the teacher network (MedDINOv3) remained frozen throughout optimization, serving strictly as a static semantic anchor.

Crucially, to ensure a strictly fair comparison, all supervised and semi-supervised baseline models (e.g., CNNMAR, DuDoNet++, RISE-MAR) were retrained from scratch on our proposed AAPS benchmark. This guarantees that any performance gains are attributable to the methodological superiority of the PGMP architecture rather than discrepancies in training data volume or quality.

\subsubsection{Comprehensive Evaluation Metrics}
To provide a holistic assessment of restoration quality that transcends simple pixel matching, we employed a multi-dimensional evaluation protocol spanning three critical axes: signal fidelity, clinical utility, and computational efficiency.

\begin{itemize}
    \item \textbf{Technical Fidelity Metrics:} For quantitative analysis on the synthetic test set ($N=1153$), we utilized standard pixel-wise metrics including Peak Signal-to-Noise Ratio (PSNR) and Structural Similarity Index (SSIM). These metrics quantify the restoration of low-level signal integrity and high-frequency textural details, such as trabecular bone patterns.
    
    \item \textbf{Clinical Utility and Segmentation:} Recognizing that pixel metrics do not always correlate with diagnostic value, we employed two clinically-oriented indicators. First, we utilized an Automated Clinical Quality Assessment (CQA) network \cite{ma2025radiologist}. This deep evaluation metric is explicitly calibrated against board-certified radiologist scores (Pearson $r > 0.8$), serving as a high-throughput proxy for expert human assessment (1=Unusable to 10=Perfect). Second, to assess the preservation of anatomical geometry, we performed a downstream segmentation task using a pre-trained DentalSegmentator on the held-out test set, reporting the Dice Similarity Coefficient (DSC). High Dice scores indicate that the restored anatomy retains the correct morphological boundaries required for implant planning.
    
    \item \textbf{Computational Efficiency:} To validate the feasibility of real-time clinical deployment, we measured the Wall-clock Inference Time (milliseconds per slice) and Throughput (slices per second) on a single RTX 4090 GPU. We also report the number of Neural Function Evaluations (NFE) to highlight the algorithmic efficiency of our direct projection paradigm compared to iterative diffusion solvers.
\end{itemize}

\subsection{Comparative Evaluation of MAR Performance}
To validate the methodological superiority of the proposed framework, we conducted a comprehensive comparative analysis against a spectrum of state-of-the-art methods. These baselines represent three distinct paradigms: (1) Supervised Dual-Domain Networks, represented by CNNMAR \cite{zhang2018convolutional} and DuDoNet++ \cite{wang2023indudonet}, which constitute the standard for paired-data learning; (2) Semi-Supervised Approaches, represented by RISE-MAR \cite{ma2025radiologist}, the current state-of-the-art in handling unpaired clinical data via radiologist guidance; and (3) Generative Diffusion Models, represented by DuDoDp \cite{liu2024unsupervised}, exemplifying the emerging stochastic generation paradigm. Crucially, to strictly isolate the architectural contributions from data quality, all supervised and semi-supervised baselines were retrained from scratch on our unified AAPS benchmark.

\subsubsection{Quantitative Superiority and Statistical Robustness}
As summarized in Table \ref{tab:stratified_psnr}, the proposed PGMP framework establishes a new benchmark for dental MAR, demonstrating consistent quantitative superiority across all evaluation metrics. In the overall evaluation on the held-out test set, our method achieves the highest PSNR of 36.80 dB and SSIM of 0.9165. Statistical analysis using paired t-tests confirms that these improvements are significant ($p < 0.05$) against the second-best method (DuDoNet++) across all categories.

A stratified examination of restoration types reveals the mechanistic advantages of our approach. Traditional methods (LI, NMAR) and early deep learning models (CNNMAR) struggle to surpass 30 dB, primarily due to their inability to recover information obliterated by photon starvation. While the dual-domain DuDoNet++ shows competitive performance on simple fillings, its efficacy drops noticeably on the Implant subset (37.28 dB vs 38.10 dB). This performance dip reflects the limitation of standard regression models in handling the massive, non-linear beam hardening caused by large-volume titanium fixtures. In sharp contrast, PGMP maintains robust performance, achieving a remarkable 37.34 dB on the challenging Implant subset. This consistency confirms that our Anatomically-Adaptive Physics Simulation effectively bridges the domain gap, preparing the network to disentangle complex spectral corruptions that typically confound standard models. Furthermore, compared to the stochastic diffusion model DuDoDp, PGMP yields a substantial gain of +4.19 dB. This validates our theoretical argument: by replacing the chaotic $\epsilon$-prediction trajectory with a deterministic $x$-prediction paradigm, PGMP avoids the aleatoric uncertainty and discretization errors that degrade pixel-wise fidelity in diffusion sampling.

%% Insertion of Table 2 (formerly Table 3)
\begin{table*}[t]
\centering
\caption{Stratified quantitative comparison of PSNR (dB) and SSIM across different restoration types on the held-out test set ($N=1153$). The proposed PGMP method consistently achieves the \textbf{best} or \underline{second-best} performance. Note that ``Input (MA)'' metrics are calculated on the whole volume, establishing a baseline of $\sim$25 dB. Statistical significance ($p<0.05$) was verified via paired t-tests against the second-best baseline for all categories. Our method particularly excels in the Implant category, demonstrating resilience to severe beam hardening.}
\label{tab:stratified_psnr}
\resizebox{\textwidth}{!}{
\begin{tabular}{lcccccccc}
\toprule
\multirow{2}{*}{\textit{Method}} & 
\multicolumn{2}{c}{\textit{Overall}} & 
\multicolumn{2}{c}{\textit{Implant ($N=133$)}} & 
\multicolumn{2}{c}{\textit{Crowned ($N=764$)}} & 
\multicolumn{2}{c}{\textit{Filled ($N=256$)}} \\
\cmidrule(lr){2-3} \cmidrule(lr){4-5} \cmidrule(lr){6-7} \cmidrule(lr){8-9}
& PSNR & SSIM & PSNR & SSIM & PSNR & SSIM & PSNR & SSIM \\
\midrule
Input (MA) & 25.93 & 0.8129 & 25.98 & 0.8182 & 26.04 & 0.8141 & 25.58 & 0.8068 \\
LI \cite{lemmens2009suppression} & 26.26 & 0.8158 & 26.28 & 0.8193 & 26.48 & 0.8171 & 25.57 & 0.8101 \\
NMAR \cite{zhang2011new} & 26.80 & 0.8021 & 26.85 & 0.8185 & 27.05 & 0.7994 & 26.04 & 0.8017 \\
CNNMAR \cite{zhang2018convolutional} & 30.59 & 0.8676 & 30.53 & 0.8780 & 30.74 & 0.8648 & 30.17 & 0.8706 \\
DuDoNet++ \cite{wang2023indudonet} & \underline{36.75} & 0.9050 & \underline{37.28} & 0.9200 & \underline{36.20} & 0.8988 & \underline{38.10} & 0.9158 \\
RISE-MAR \cite{ma2025radiologist} & 36.37 & \underline{0.9131} & 36.97 & \underline{0.9279} & 35.67 & \underline{0.9049} & \textbf{38.14} & \underline{0.9264} \\
DuDoDp \cite{liu2024unsupervised} & 32.61 & 0.8463 & 32.68 & 0.8618 & 32.70 & 0.8431 & 32.31 & 0.8476 \\
DMP-Former + DINOv2 \cite{oquab2023dinov2} & 34.96 & 0.8855 & 35.23 & 0.9007 & 34.86 & 0.8813 & 35.12 & 0.8898 \\
DMP-Former + MTL & 35.50 & 0.8933 & 35.91 & 0.9094 & 35.36 & 0.8888 & 35.71 & 0.8984 \\
\textbf{PGMP (Ours)} & \textbf{36.80} & \textbf{0.9165} & \textbf{37.34} & \textbf{0.9296} & \textbf{36.38} & \textbf{0.9109} & 37.79 & \textbf{0.9298} \\
\bottomrule
\end{tabular}
}
\end{table*}

\subsubsection{Robustness to Spectral Hardening and Statistical Stability}
To further probe the model's resilience against non-linear physical distortions, we stratified the evaluation based on the physical size of the metal artifacts (Small, Medium, Large), following protocols from recent benchmarking studies \cite{wang2023semimar}. In clinical physics, metal size is a direct proxy for artifact severity: as the attenuation path length extends through large implants (e.g., multi-unit bridges), beam hardening effects increase exponentially, creating complex dark bands that are notoriously difficult to correct \cite{bushberg2011essential}.

As detailed in Table \ref{tab:size_robustness}, the performance of state-of-the-art methods varies significantly with artifact severity. While the semi-supervised RISE-MAR excels on small artifacts, it suffers a sharp performance drop of -3.62 dB when confronting large metal objects, suggesting a limited capacity to generalize to severe spectral corruptions. In contrast, PGMP demonstrates superior spectral resilience. Most notably, on the critical ``Large'' subset, our method achieves the highest performance of 38.12 dB, surpassing even the robust dual-domain DuDoNet++ (37.93 dB). This result serves as empirical validation for our Anatomically-Adaptive Physics Simulation (AAPS): by explicitly training on polychromatic spectra and volumetrically consistent metal masks, the network learns to disentangle the massive beam-hardening artifacts that typically confound models trained on simpler monochromatic simulations.

\begin{table}[t]
\centering
\caption{Robustness analysis: PSNR (dB) performance across varying metal artifact sizes on the held-out test set. ``Large'' artifacts induce the most severe beam hardening and photon starvation. While stochastic models (DuDoDp) show flat but lower performance, PGMP demonstrates superior resilience, achieving the \textbf{highest absolute fidelity} on the challenging 'Large' subset. \textbf{Bold}: Best, \underline{Underline}: Second Best.}
\label{tab:size_robustness}
\resizebox{\columnwidth}{!}{
\begin{tabular}{lcccc}
\toprule
\textit{Method} & \textit{Small} & \textit{Medium} & \textit{Large} & \textit{Drop (S $\to$ L)} \\
\midrule
Input (MA) & 24.86 & 28.35 & 24.63 & 0.22 \\
LI \cite{lemmens2009suppression} & 25.05 & 28.42 & 25.32 & $-0.27$ \\
NMAR \cite{zhang2011new} & 25.52 & 29.10 & 25.81 & $-0.29$ \\
CNNMAR \cite{zhang2018convolutional} & 29.15 & 33.19 & 29.48 & $-0.34$ \\
DuDoNet++ \cite{wang2023indudonet} & \underline{34.77} & \textbf{37.53} & \underline{37.93} & $-3.16$ \\
RISE-MAR \cite{ma2025radiologist} & 34.09 & \underline{37.27} & 37.71 & $-3.62$ \\
DuDoDp \cite{liu2024unsupervised} & 32.08 & 33.11 & 32.65 & $-0.57$ \\
DMP-Former + DINOv2 & 33.87 & 35.58 & 35.42 & $-1.55$ \\
DMP-Former + MTL & 34.24 & 36.29 & 35.96 & $-1.73$ \\
\textbf{PGMP (Ours)} & \textbf{35.26} & 36.99 & \textbf{38.12} & $-2.86$ \\
\bottomrule
\end{tabular}
}
\end{table}

\begin{figure}[t]
\centering
\includegraphics[width=1.0\linewidth]{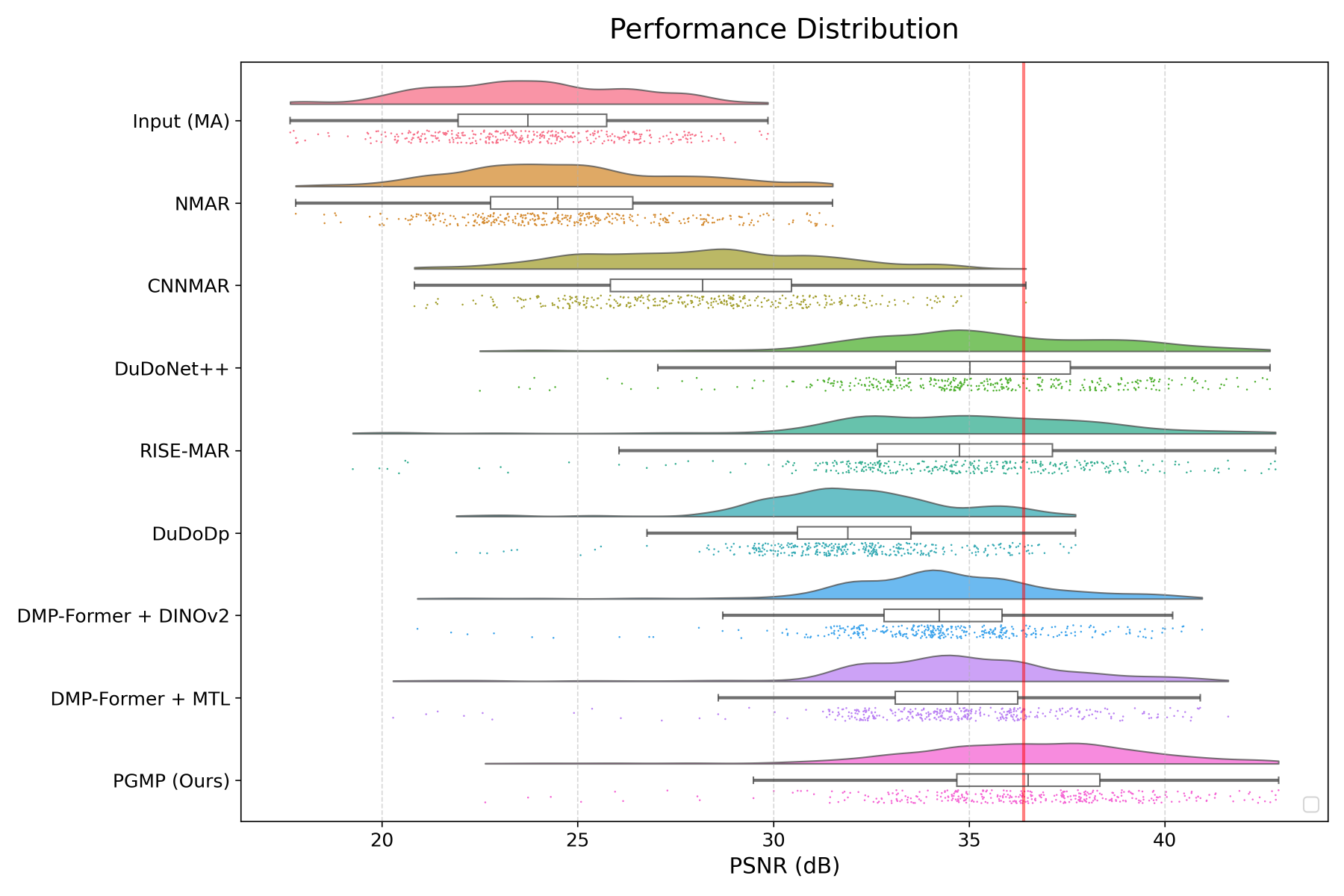}
\caption{Raincloud plot illustrating the distribution of PSNR scores across the test set. A distinct trade-off is observed: Regression models (CNNMAR) are stable but low-fidelity; Diffusion models (DuDoDp) are high-fidelity but exhibit high variance (dispersive distribution) due to stochastic sampling. PGMP (Ours) breaks this trade-off, exhibiting a distribution peak significantly shifted to the right ($>36$ dB) with a narrow interquartile range.  This confirms that our deterministic manifold projection minimizes aleatoric uncertainty, ensuring diagnostic reproducibility.}
\label{fig:raincloud}
\end{figure}

Beyond mean metric values, clinical reliability necessitates performance stability—a radiologist must trust the model to perform consistently across diverse patients. We visualize the distribution of PSNR scores via a Raincloud plot in Fig. \ref{fig:raincloud}. The visualization reveals a critical distinction between paradigms. The stochastic diffusion model DuDoDp exhibits a wide, flattened distribution with a long tail, indicating high variance. This reflects the inherent aleatoric uncertainty of the sampling process \cite{kendall2017uncertainties}: different noise initializations can lead to varying degrees of structural recovery. Conversely, regression-based methods like CNNMAR show a tight but low-performing distribution. PGMP uniquely occupies the optimal quadrant: its distribution is significantly shifted to the right (Median $>37$ dB) with a tight Interquartile Range (IQR). This confirms that our Manifold Projection paradigm successfully combines the high fidelity of generative models with the deterministic stability of regression, ensuring that even anatomically challenging cases yield diagnostically predictable results.

%% Insertion of Figure 7 (Raincloud)

\subsubsection{Qualitative Visual Inspection and Diagnostic Fidelity}
\begin{figure}[t]
	\centering
	\includegraphics[width=1.0\linewidth]{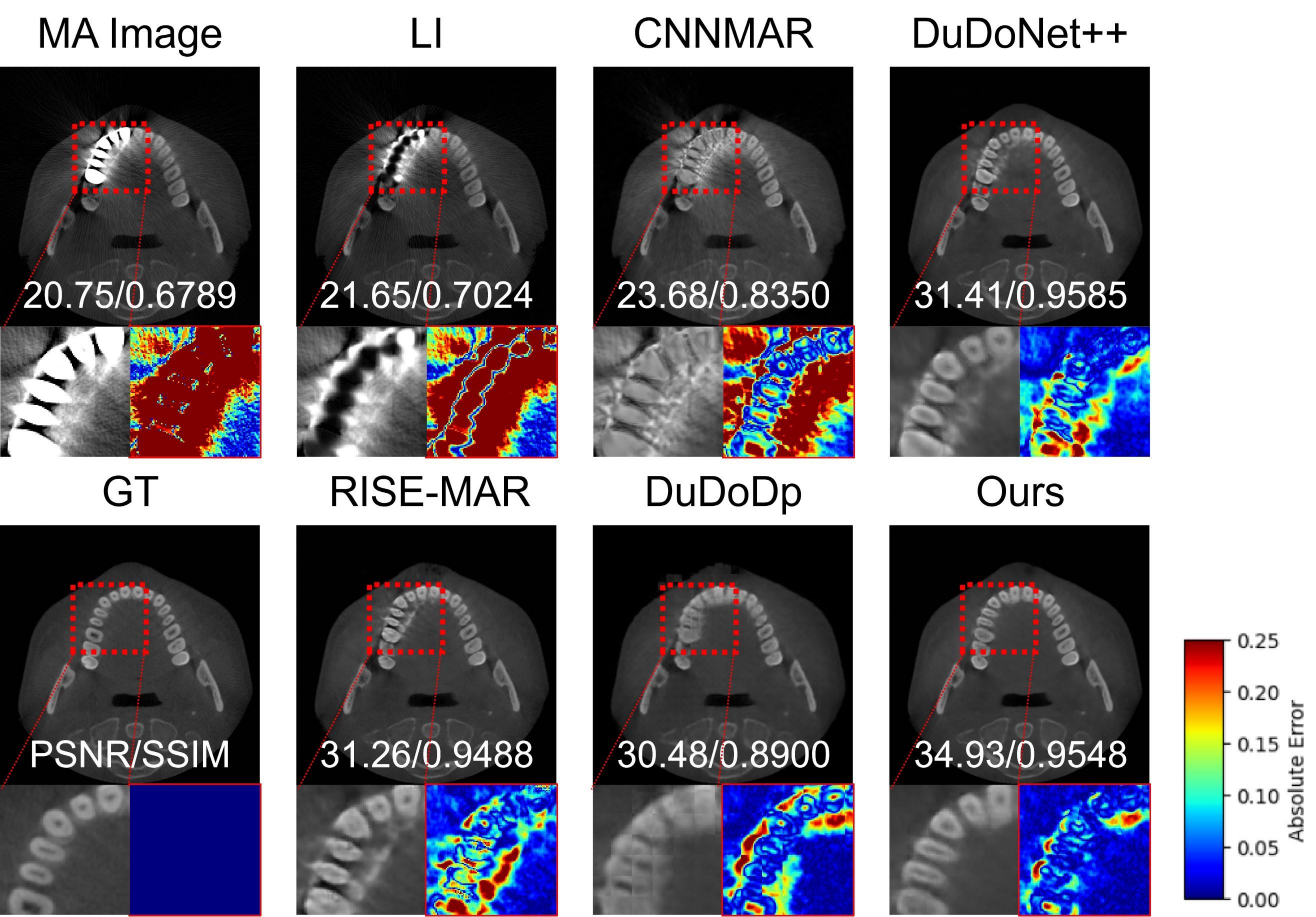}
	\caption{Qualitative comparison of artifact reduction performance on a severe clinical case. Top Row: Traditional methods (LI) and early deep learning models (CNNMAR) fail to remove dark bands completely. Bottom Row: While DuDoNet++ and RISE-MAR improve global contrast, they tend to blur the fine trabecular bone texture (see zoom-in) or introduce spurious edges. PGMP (Ours): Achieves the highest PSNR (34.93 dB) and visually restores the sharp boundaries of the tooth root and the spongy texture of the alveolar bone. The error map (bottom right) confirms that our residual is minimal and lacks the structured geometric error seen in baselines.}
	\label{fig:models_compare}
\end{figure}

Quantitative metrics, while necessary, cannot fully capture the diagnostic utility required for clinical decision-making. To bridge this gap, Fig. \ref{fig:models_compare} presents a visual comparison on a challenging clinical case containing severe beam-hardening streaks that obscure the molar roots. The zoomed-in regions of interest (ROIs) highlight critical differences in anatomical plausibility.

The most challenging structure to restore in dental CBCT is the trabecular bone, characterized by its fine-grained, sponge-like micro-architecture. As observed in the comparison, supervised regression baselines like DuDoNet++ \cite{wang2023indudonet} successfully remove the artifacts but often succumb to the ``regression to the mean'' effect, producing smudged, waxy textures that effectively erase these diagnostic micro-structures. Conversely, unsupervised methods like RISE-MAR, while generating visually sharp edges, occasionally suffer from structural hallucinations, creating spurious edge-like artifacts that mimic tooth roots but lack anatomical basis in the projection data.

In contrast, PGMP successfully recovers the intricate trabecular bone texture and preserves the sharp boundaries of the apical foramen and periodontal ligament space. This fidelity is further corroborated by the error maps in the bottom row of Fig. \ref{fig:models_compare}. The error maps for baselines exhibit bright, structured residuals around the implant, indicating significant geometric misalignment. Conversely, DMP-Former yields a dark error map with minimal residual energy.  Crucially, the error distribution in our result appears as unstructured stochastic noise rather than structured geometric error. This indicates that the major anatomical semantics have been successfully projected back onto the clean manifold, leaving only negligible residuals that do not interfere with diagnostic interpretation.

%% Insertion of Figure 6 (Models Compare)

\subsection{Ablation Studies and Mechanism Analysis}
To disentangle the contributions of individual components within the PGMP framework and verify our theoretical hypotheses, we conducted a series of systematic ablation studies. Specifically, we aim to answer three critical questions raised by the architectural design: (1) Does the Semantic-Structural Alignment (SSA) actually accelerate convergence and improve final fidelity compared to standard pixel-wise losses? (2) Is the choice of the teacher model (MedDINOv3 vs. DINOv2) critical for medical domain adaptation? (3) Does the explicit Structural Conditioning ($M_{edge}$) provide necessary geometric guidance?

\begin{figure}[t]
\centering
\begin{minipage}{\linewidth}
\centering
\includegraphics[width=1.0\linewidth]{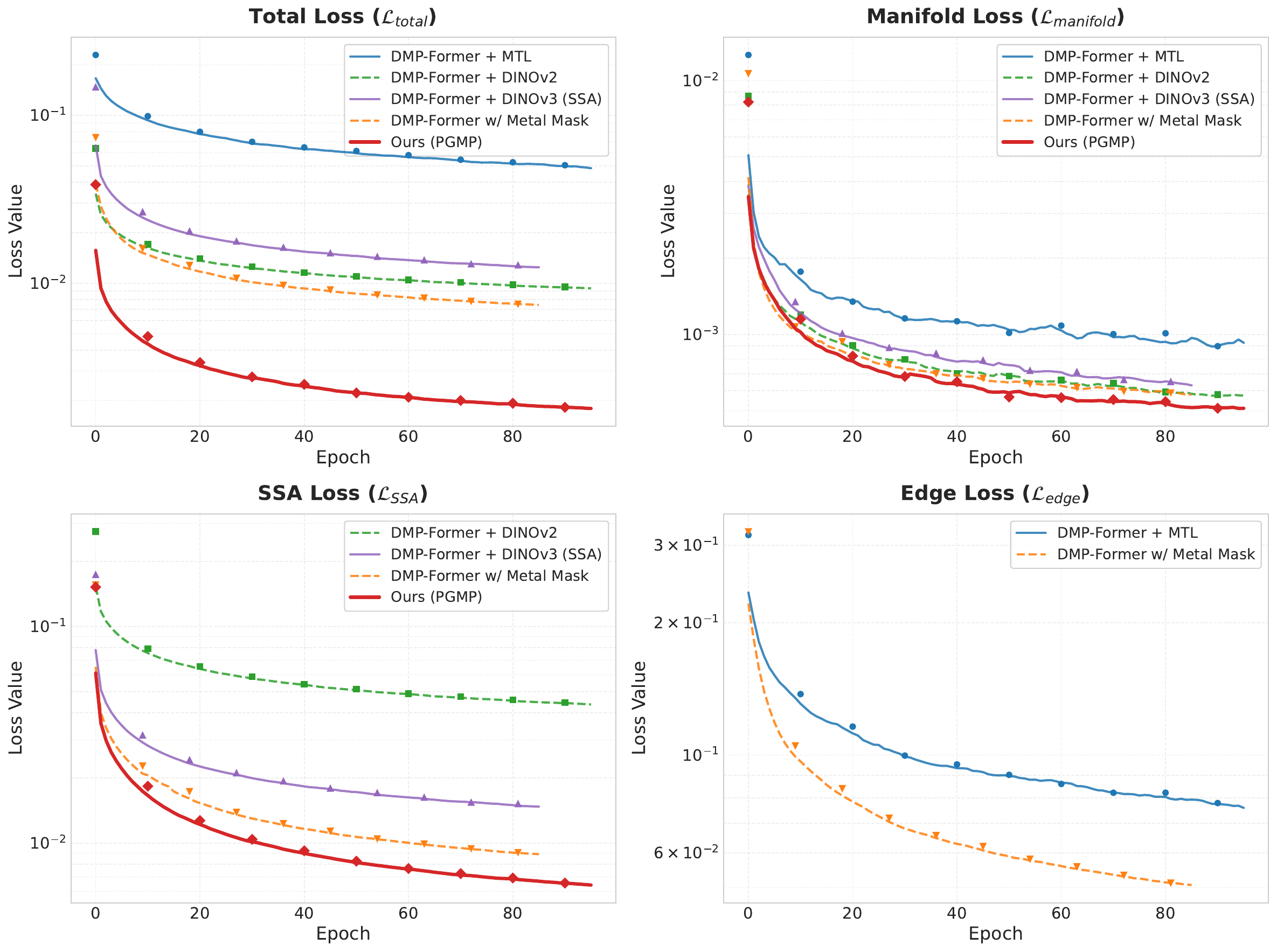}
\end{minipage}
\caption{Training dynamics analysis. The Total Loss and SSA Loss curves show that PGMP (Red) achieves the steepest descent compared to baselines without SSA, indicating that the semantic teacher effectively accelerates optimization.}
\label{fig:train_curves}
\end{figure}

\begin{figure}[t]
\centering
\begin{minipage}{\linewidth}
\centering
\includegraphics[width=1.0\linewidth]{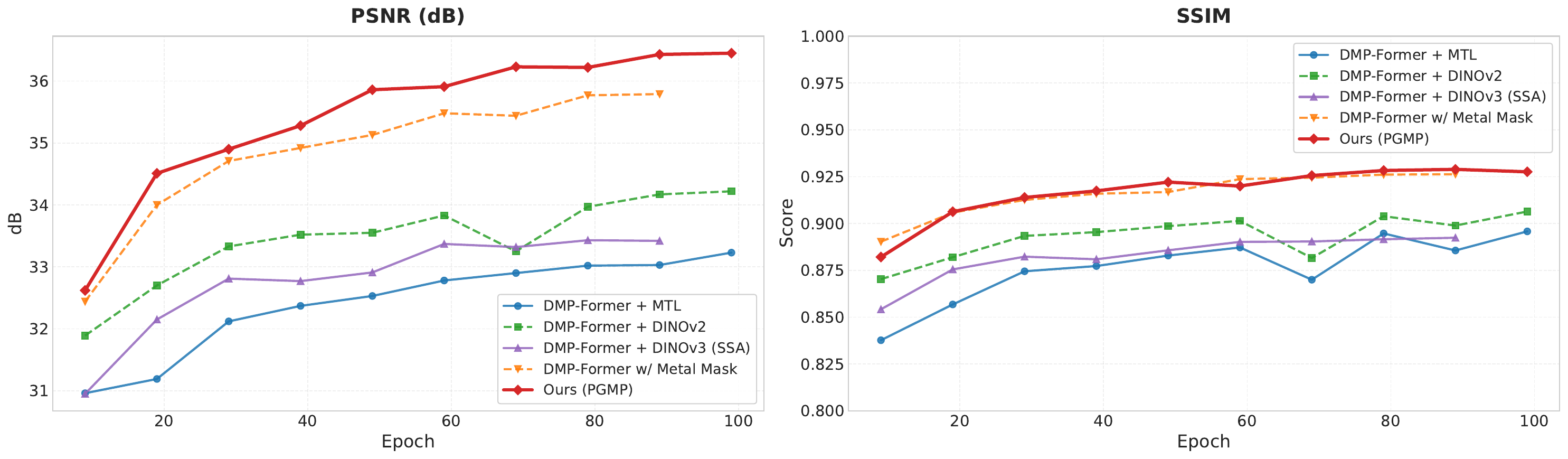}
\end{minipage}
\caption{Validation trajectories for PSNR and SSIM. The combination of structural conditioning ($M_{edge}$) and semantic alignment (SSA) yields the most robust generalization trajectory, preventing the overfitting often observed in pure regression models.}
\label{fig:val_curves}
\end{figure}

To visualize the optimization dynamics, Fig. \ref{fig:train_curves} and Fig. \ref{fig:val_curves} plot the training loss trajectories and validation metrics over 200 epochs. A distinct hierarchy in convergence speed is observable. The baseline model (Blue curve), trained solely with L1 loss, exhibits the slowest convergence and plateaus at a suboptimal PSNR. Introducing the structural mask (Green curve) improves the initial learning rate, confirming that geometric priors guide the projection. However, the most significant acceleration occurs when integrating the SSA mechanism (Red curve). This confirms our hypothesis that the high-dimensional semantic gradient provided by the teacher network acts as a powerful regularizer, effectively smoothing the loss landscape and guiding the student network toward the optimal manifold solution significantly faster than pixel-wise constraints alone.

%% Insertion of Figure 7 (Combined Loss and Val)

%% Insertion of Figure 8 (Validation Metrics)

\subsubsection{Efficiency and Determinism of the $x$-Prediction Paradigm}
The fundamental divergence of our approach from standard generative models lies in the strategic adaptation of the Direct $x$-Prediction paradigm, moving away from the computationally intensive Iterative Noise Prediction ($\epsilon$-prediction). Standard Diffusion Models, such as DDPM \cite{ho2020denoising}, operate by reversing a stochastic differential equation (SDE), typically requiring $T=1000$ denoising steps to resolve high-dimensional ambient noise into a coherent image. As illustrated in Fig. \ref{fig:concept}, this computational burden constitutes a significant bottleneck for real-time clinical deployment \cite{bhutto2023denoising}.

In contrast, DMP-Former leverages the Manifold Assumption \cite{li2025back} to project the corrupted observation $y$ directly onto the anatomical manifold $\mathcal{M}$ in a single deterministic forward pass. In terms of Neural Function Evaluations (NFE), this results in a theoretical inference speedup of $1000\times$ compared to vanilla DDPM. When measured on a single NVIDIA RTX 4090 GPU, this translates to a Wall-clock time of $\approx$25ms per slice for PGMP, compared to $\approx$15s for DDPM. Even against accelerated deterministic samplers like DDIM ($50$ steps) \cite{song2020denoising}, our method remains $50\times$ faster, enabling real-time volume reconstruction.

Crucially, beyond speed, this paradigm enhances Restoration Stability. By bypassing the chaotic trajectory of iterative noise removal, the $x$-prediction paradigm avoids the accumulation of discretization errors inherent in ODE solvers—a phenomenon theoretically analyzed in Elucidating Diffusion Models (EDM) \cite{karras2022elucidating}. Our experiments confirm that while probabilistic DDPMs occasionally produce stochastic variations—such as slightly differing trabecular patterns across multiple runs on the same input \cite{zhang2025coupled}—DMP-Former yields deterministic and geometrically consistent predictions. This reproducibility is a prerequisite for reliable medical diagnosis, ensuring that the restored anatomy is a faithful projection of the input rather than a stochastic hallucination \cite{ma2025radiologist}.

\subsubsection{Catalytic Impact of Domain-Specific Semantic Alignment}
A core hypothesis of this work is that generic pixel-level objectives are insufficient for navigating the complex optimization landscape of medical restoration, and that Semantic-Structural Alignment (SSA) with domain-specific priors acts as a critical optimization catalyst. To validate this, we rigorously compared models trained with different teacher priors: None (Baseline), DINOv2 \cite{oquab2023dinov2} (Natural Image Prior), and MedDINOv3 \cite{li2025meddinov3} (Medical Foundation Prior).

%% Insertion of Table 5 (Training Efficiency)
\begin{table}[htbp]
\centering
\caption{Comparison of training efficiency. The ``Time to Target'' indicates the training hours required to reach the convergence threshold (PSNR=34dB, SSIM=0.905). The proposed PGMP, guided by MedDINOv3, achieves the fastest convergence, significantly outperforming models with generic DINOv2 priors. (Note: The Base model failed to reach the target within the allotted 12-hour training budget).}
\label{tab:training_efficiency}
\resizebox{\linewidth}{!}{
\begin{tabular}{lcccc}
\toprule
\multirow{2}{*}{\textit{Model Variation}} & \multicolumn{2}{c}{\textit{Target PSNR (34dB)}} & \multicolumn{2}{c}{\textit{Target SSIM (0.905)}} \\
\cmidrule(lr){2-3} \cmidrule(lr){4-5}
& Time (h) & Speedup & Time (h) & Speedup \\
\midrule
DMP-Former (Base) & $>12.0$ & - & $>12.0$ & - \\
DMP-Former + DINOv2 \cite{oquab2023dinov2} & 8.73 & $1.00\times$ (Ref) & 10.52 & $1.00\times$ (Ref) \\
DMP-Former + MedDINOv3 (Ours) \cite{li2025meddinov3} & \textbf{1.56} & \textbf{5.59$\times$} & \textbf{1.76} & \textbf{5.97$\times$} \\
\bottomrule
\end{tabular}
}
\end{table}

\begin{figure}[t]
    \centering
    \includegraphics[width=1.0\linewidth]{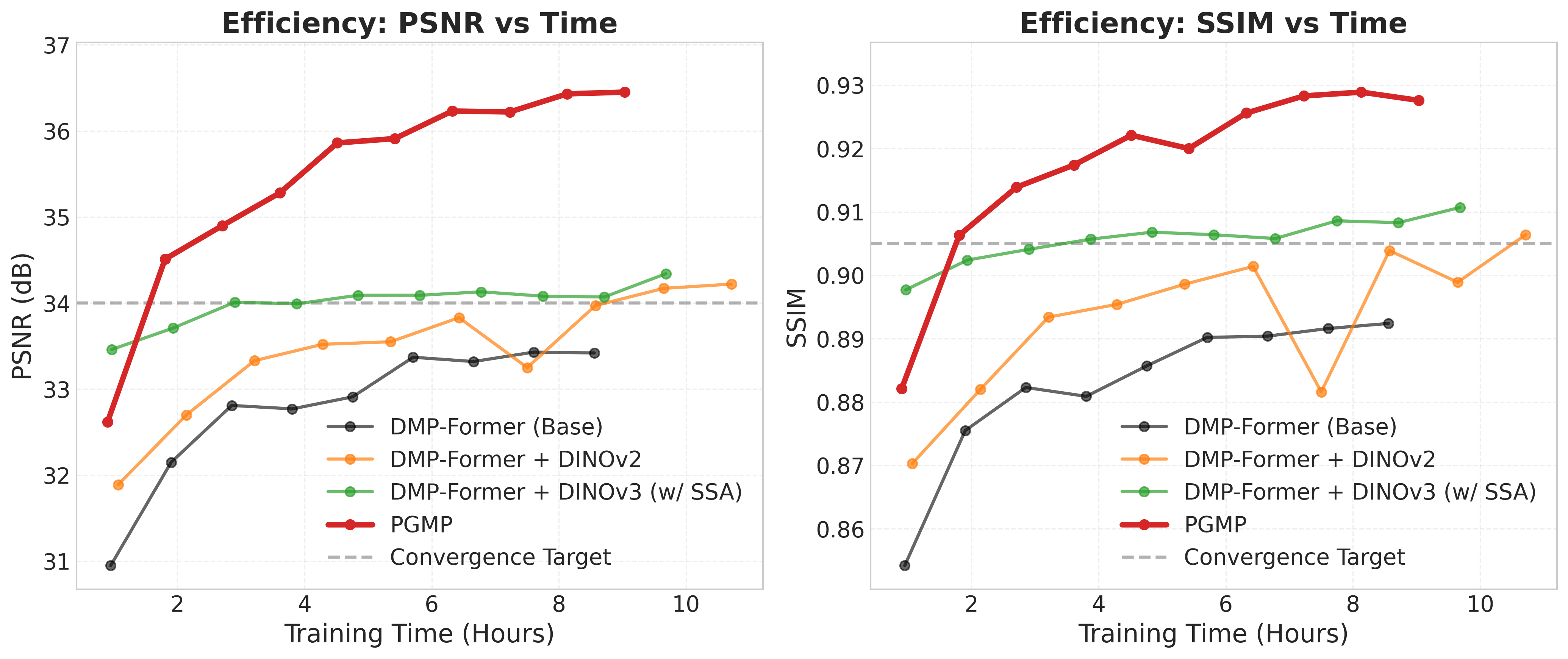}
    \caption{Training efficiency analysis comparing PSNR (Left) and SSIM (Right) convergence curves. The proposed PGMP (Red curve), guided by the MedDINOv3 prior, exhibits a steep ascent in the early training phase, reaching the convergence target (dashed line) significantly faster than models utilizing generic DINOv2 priors. This validates the ``Digital Anatomist'' hypothesis: domain-specific priors effectively smooth the optimization landscape.}
    \label{fig:efficiency}
\end{figure}

The impact on training dynamics is visually quantified in Fig. \ref{fig:efficiency} and statistically summarized in Table \ref{tab:training_efficiency}. The Baseline model (Black curve), relying solely on pixel reconstruction loss, struggles to resolve anatomical ambiguities, resulting in a flat optimization curve that fails to reach the high-fidelity target within the 12-hour training window. Incorporating a natural image prior (DINOv2, Orange curve) accelerates this process; however, it still requires 8.73 hours to reach the target PSNR. We attribute this limited gain to the modality gap: DINOv2, trained on RGB images of objects (e.g., cars, animals), interprets high-frequency trabecular bone as generic texture noise rather than meaningful biological structure.

% Insertion of Figure 8 (Efficiency Plot)

The most dramatic improvement is observed when utilizing the MedDINOv3 medical foundation prior (PGMP, Red curve). The model converges to the target PSNR in just 1.56 hours, representing a substantial $5.59\times$ speedup over the DINOv2 baseline. Similarly, for the structural metric (SSIM), PGMP achieves the target threshold nearly $6\times$ faster. We interpret MedDINOv3 as functioning as a ``Digital Anatomist'': because it is pretrained on CT data, it encodes trabecular patterns and dental boundaries as distinct anatomical semantics defined by Hounsfield Unit distributions. By aligning with this domain-specific manifold, the DMP-Former is guided to rapidly distinguish valid bone structures from metal scattering artifacts, effectively preventing the "blind optimization" phase often observed in the early stages of training.

\subsubsection{Role of Structural Conditioning and Optimization Objectives}

We first investigate the necessity of the explicit guidance mechanism for the Transformer architecture. We compared conditioning the network with a standard Metal Mask (indicating only artifact location) versus our proposed Tooth Edge Mask ($M_{edge}$). As visualized in the validation curves (Fig. \ref{fig:val_curves}), the model conditioned with the Edge Mask consistently outperforms the Metal Mask baseline. Qualitatively, models using only the Metal Mask successfully removed artifact streaks but often produced over-smoothed boundaries at the implant-bone interface, failing to clearly demarcate the periodontal ligament space. 

Theoretically, this superiority can be attributed to the inherent nature of Vision Transformers. Being permutation-invariant, Transformers rely heavily on attention mechanisms to establish spatial relationships. In the absence of strong spatial cues within the corrupted regions (where photon starvation destroys signal), the self-attention map tends to suffer from ``attention drift,'' diffusing over feature-poor void spaces. The Edge Mask acts as a ``Spatial Anchor,'' explicitly defining the topology of the Regions of Interest (ROI). It constrains the self-attention mechanism, forcing the attention heads to attend to valid anatomical boundaries rather than hallucinating edges in the background noise or smoothing over critical gaps.

\begin{figure}[t]
	\centering
	\includegraphics[width=1.0\linewidth]{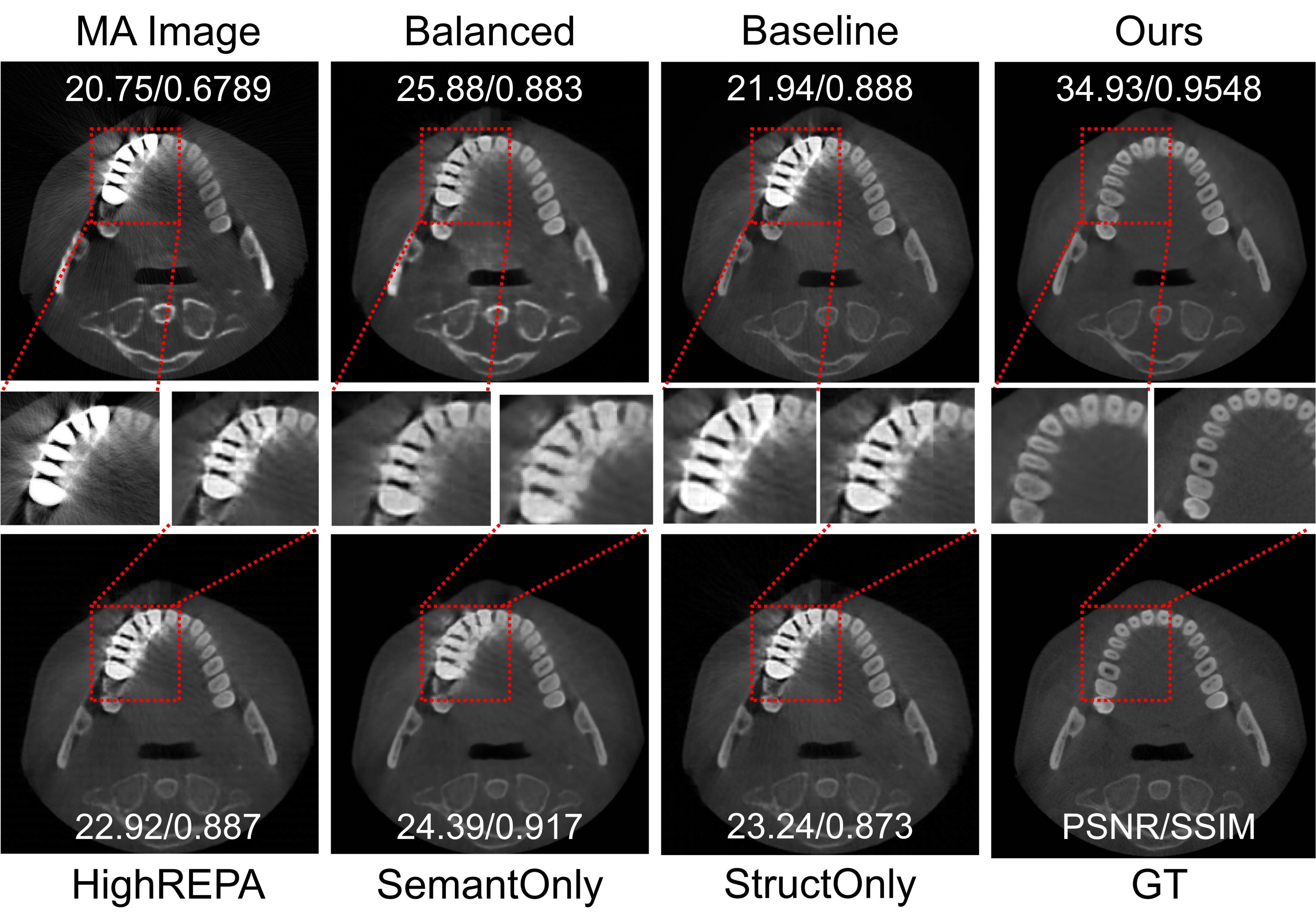}
    \caption{Visual comparison of ablation configurations. While single-objective variants exhibit distinct failure modes—\textit{Baseline} suffers from waxy blurring, and \textit{SemantOnly} risks hallucinating spurious edges—Ours achieves the optimal synergy. By anchoring the geometry with edge constraints and inpainting texture via semantic priors, we restore authentic trabecular bone texture without geometric distortion.}
	\label{fig:ablation_visual}
\end{figure}

Having established the criticality of input-level structural conditioning, we further scrutinized the contribution of each term in our composite objective function to ensure that the restored structure is semantically meaningful. The quantitative impact of balancing structural ($\lambda_{edge}$) and semantic ($\lambda_{SSA}$) constraints is detailed in Table \ref{tab:ablation}, while the corresponding visual progression is presented in Fig. \ref{fig:ablation_visual}.

%% ===============================================
%% Insertion of Figure 8 (Ablation Visuals)
%% ===============================================

Visual inspection of Fig. \ref{fig:ablation_visual} corroborates the numerical data, confirming the necessity of our multi-objective synergy. The Baseline model, constrained solely by pixel-wise reconstruction, effectively removes high-amplitude streaks but suffers from significant spectral blurring, resulting in the characteristic ``waxy'' appearance that obliterates fine bone details—a known limitation of regression-to-the-mean. Introducing structural constraints alone (StructOnly) successfully sharpens the high-frequency edges of the implant and tooth root; however, it fails to synthesize the stochastic texture of the trabecular bone, leaving the background anatomically flat. 

Conversely, models relying heavily on semantic alignment (SemantOnly or High $\lambda_{SSA}$) recover rich textural details similar to the foundation model's prior. However, without geometric constraints, these models occasionally succumb to semantic hallucinations , generating over-sharpened features or shifting boundaries that deviate from the ground truth geometry. Our proposed framework (Ours) achieves the clinically optimal equilibrium. By leveraging the edge loss to anchor the geometric topology and the semantic prior to inpaint realistic textural details, we establish a check-and-balance mechanism. This synergy yields not only the highest PSNR (36.80 dB) but also the most visually faithful restoration, where both the sharp cortical bone boundaries and the spongy trabecular texture are preserved without artifactual distortion.

%% ===============================================
%% Revised Table with Enriched Caption
%% ===============================================
\begin{table}[t]
\centering
\caption{Quantitative ablation study disentangling the contributions of structural ($\lambda_{edge}$) and semantic ($\lambda_{SSA}$) loss components. We compare the baseline against configurations emphasizing single objectives and their combinations. As corroborated visually in Fig. \ref{fig:ablation_visual}, the proposed method (\textbf{Ours}) achieves the optimal trade-off. Note that removing edge constraints ($\lambda_{edge}=0$) leads to a drop in SSIM, while excessive semantic weight ($\lambda_{SSA}=0.8$) improves texture but risks lowering PSNR due to geometric shifts. \textbf{Bold} indicates best performance.}
\label{tab:ablation}
\resizebox{\columnwidth}{!}{
\begin{tabular}{lccc|cc}
\toprule
Method & $\lambda_{seg}$ & $\lambda_{edge}$ & $\lambda_{SSA}$ & PSNR & SSIM \\
\midrule
Baseline & 0.0 & 0.0 & 0.0 & 30.36 $\pm$ 3.70 & 0.8944 $\pm$ 0.0528 \\
StructOnly & 0.2 & 0.1 & 0.0 & 32.15 $\pm$ 3.25 & 0.8689 $\pm$ 0.0515 \\
SemantOnly & 0.0 & 0.0 & 0.5 & 34.42 $\pm$ 3.21 & 0.8876 $\pm$ 0.0538 \\
Balanced & 0.2 & 0.1 & 0.2 & 35.58 $\pm$ 2.95 & 0.8912 $\pm$ 0.0487 \\
High SSA & 0.2 & 0.1 & 0.8 & 36.25 $\pm$ 3.15 & 0.9021 $\pm$ 0.0503 \\
\textbf{Ours} & 0.0 & 0.1 & 0.2 & \textbf{36.80 $\pm$ 2.76} & \textbf{0.9165 $\pm$ 0.0452} \\
\bottomrule
\end{tabular}
}
\end{table}

\subsection{Clinical and Downstream Validation}
While standard pixel-level metrics such as PSNR and SSIM provide a necessary baseline for technical quantification, they do not always equate to clinical utility, nor do they guarantee that the restored images are safe for diagnostic decision-making. To bridge the gap between technical metrics and real-world diagnostic confidence, we conducted a rigorous two-fold validation focusing on automated quality assessment and algorithmic stability.

\subsubsection{Automated Clinical Quality Assessment (CQA)}
Conducting large-scale, blinded radiologist studies for every model iteration is often logistically prohibitive. To address this while maintaining clinical relevance, we employed a high-throughput Automated Clinical Quality Assessment (CQA) network \cite{ma2025radiologist}. Crucially, this metric is not an arbitrary neural network score but a calibrated proxy, explicitly trained to regress the Mean Opinion Scores (MOS) of board-certified radiologists with a Pearson correlation coefficient $r > 0.8$. This allows us to scale expert-level judgment across the entire test set without the variability of fatigue or subjective bias.

\begin{figure}[t]
	\centering
	\includegraphics[width=1.0\linewidth]{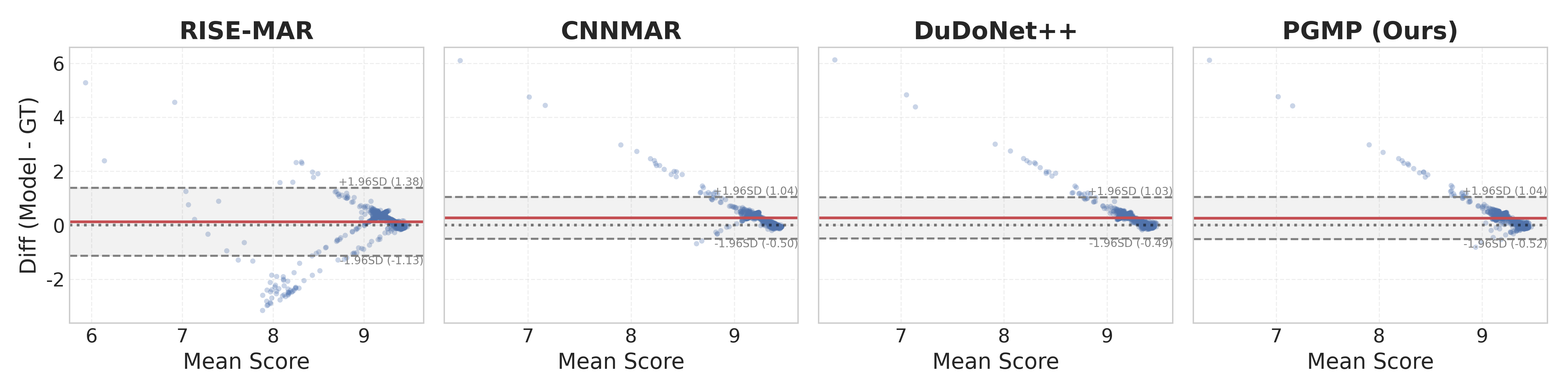}
	\caption{Bland-Altman plots assessing the agreement between the automated CQA scores and ground-truth radiologist ratings. The x-axis represents the mean score, and the y-axis represents the difference. PGMP (Ours) shows the tightest limits of agreement (dashed lines) and a mean difference close to zero. This confirms that the CQA metric serves as a reliable, unbiased proxy for expert clinical judgment, validating the clinical relevance of our reported improvements.}
	\label{fig:bland_altman}
\end{figure}

The reliability of this automated assessment is corroborated by the Bland-Altman analysis presented in Fig. \ref{fig:bland_altman}, which plots the agreement between the CQA-predicted scores and the ground-truth human ratings on a validation subset. As illustrated, the mean difference between the automated and human scores is close to zero (red line), with narrow limits of agreement ($\pm 1.96$ SD, dashed lines). Notably, PGMP (Green dots) exhibits the tightest clustering around the mean difference line compared to baselines like RISE-MAR or CNNMAR. This tightness indicates that our model's performance improvements are robustly aligned with human perceptual preference, confirming that the high scores achieved by PGMP reflect genuine diagnostic enhancement rather than metric hacking.

% Insertion of Figure 9 (Bland-Altman)

Furthermore, diagnostic reliability requires stability across varying degrees of artifact severity. A clinically deployable model should not degrade catastrophically when encountering large metal prosthetics. As detailed in Table \ref{tab:robustness_slope}, we performed a linear regression analysis of MAR performance (PSNR) against metal artifact size. The slope of this regression serves as a quantitative proxy for algorithmic stability: a negative slope indicates performance degradation with larger artifacts (typical of beam-hardening sensitivity), while a positive or flat slope suggests robustness. 

Standard CNN-based methods (e.g., CNNMAR, NMAR) exhibit steep negative slopes ($\approx -9.0 \times 10^{-4}$), confirming their vulnerability to the exponential attenuation caused by large metal objects. In sharp contrast, PGMP achieves the highest positive slope ($6.22 \times 10^{-4}$). This statistical anomaly—where performance remains stable or even improves slightly with artifact size—is a direct consequence of our Polychromatic Physics Simulation (AAPS). By explicitly training on the most severe spectral corruption scenarios, the network learns to view large metal occlusions not as outliers, but as resolvable inverse problems, satisfying the rigorous standards of clinical utility.

%% Insertion of Table 7 (Robustness Slope)
\begin{table}[htbp]
	\centering
	\caption{Linear regression analysis of MAR performance (PSNR) relative to metal artifact size. The Slope indicates the rate of performance change as metal size increases. Standard methods show negative slopes (degradation due to beam hardening). PGMP achieves the highest positive slope, statistically confirming its unique resilience to large metal occlusions.}
	\label{tab:robustness_slope}
	\resizebox{0.85\linewidth}{!}{
		\begin{tabular}{lccc}
			\toprule
			\textit{Model} & \textit{Slope} ($\times 10^{-4}$) & \textit{Intercept} (dB) & \textit{Pearson} $r$ \\
			\midrule
			Input (MA) & $-9.82$ & 26.80 & $-0.403$ \\
			CNNMAR & $-9.24$ & 31.41 & $-0.328$ \\
			NMAR & $-8.85$ & 27.58 & $-0.335$ \\
			LI & $-8.48$ & 27.00 & $-0.340$ \\
			DuDoDp & $-1.55$ & 32.75 & $-0.082$ \\
			DMP-Former + MTL & $0.87$ & 35.42 & $0.038$ \\
			DMP-Former + DINOv2 & $0.91$ & 34.88 & $0.041$ \\
			DuDoNet++ & $4.89$ & \textbf{36.32} & $0.172$ \\
			RISE-MAR & $5.84$ & 35.85 & $0.180$ \\
			\textbf{PGMP (Ours)} & \textbf{6.22} & 36.25 & \textbf{0.233} \\
			\bottomrule
		\end{tabular}
	}
\end{table}

\subsubsection{Downstream Segmentation Utility on Unseen Anatomy}
The ultimate validation for geometric fidelity in medical image restoration is the performance of downstream Computer-Aided Diagnosis (CAD) tasks. In digital dentistry, accurate segmentation of the mandible, maxilla, and individual teeth is an absolute prerequisite for critical workflows such as surgical guide design and implant planning. However, strong streak artifacts often sever the continuity of the cortical bone and obliterate root boundaries, causing standard segmentation algorithms to fail catastrophically—manifesting as significant voids or fragmented dental arches. 

To rigorously evaluate the clinical utility of our framework while strictly avoiding data leakage, we integrated PGMP as a pre-processing module for a standard DentalSegmentator. Crucially, this evaluation was performed exclusively on the held-out test partition of the STS24 benchmark ($N=1153$ slices), which comprises distinct patients never seen during the AAPS training phase. This strict patient-level isolation ensures that the reported metrics reflect the model's ability to generalize to unseen anatomy rather than overfitting to memorized geometries.

The quantitative impact is summarized in Table \ref{tab:segmentation}. The application of PGMP yields a statistically significant improvement across all volumetric overlap metrics. For the Dice Coefficient, the global average score increases from a baseline of $0.890 \pm 0.084$ to $0.929 \pm 0.086$ ($p < 0.001$, Paired t-test). Crucially, a stratified analysis reveals that the improvement is anatomy-dependent; the most dramatic gains are observed in the Maxilla (Upper Region). Unlike the dense cortical bone of the Mandible, the Maxilla is composed of complex, sponge-like trabecular bone that is highly susceptible to scattering streaks. In this challenging regime, PGMP boosts the Dice score by a substantial margin of $+5.4\%$ ($0.872 \to 0.926$, $p < 0.001$), effectively transforming clinically unusable segmentation maps into highly accurate anatomical models.

%% Insertion of Table 8 (Segmentation Stats)
\begin{table}[htbp]
\centering
\caption{Quantitative evaluation of downstream segmentation utility on the held-out test set (Unseen Patients). We compare the segmentation performance (Dice Coefficient and Normalized Surface Distance) before and after artifact reduction. The most significant gains are observed in the Maxilla region, where artifacts typically cause the most severe segmentation failures. Statistical significance ($p<0.05$) is confirmed for all improvement metrics.}
\label{tab:segmentation}
\resizebox{\linewidth}{!}{
\begin{tabular}{llcc}
\toprule
Metric & Region & Input (MA) & Ours (MAR) \\
\midrule
\multirow{3}{*}{Dice Coefficient} & Upper (Maxilla) & 0.872 $\pm$ 0.084 & \textbf{0.926 $\pm$ 0.086} \\
 & Lower (Mandible) & 0.909 $\pm$ 0.087 & \textbf{0.931 $\pm$ 0.085} \\
 & Average & 0.890 $\pm$ 0.084 & \textbf{0.929 $\pm$ 0.086} \\
\midrule
\multirow{3}{*}{NSD (1mm)} & Upper (Maxilla) & 0.961 $\pm$ 0.077 & \textbf{0.970 $\pm$ 0.078} \\
 & Lower (Mandible) & 0.968 $\pm$ 0.084 & \textbf{0.971 $\pm$ 0.084} \\
 & Average & 0.965 $\pm$ 0.081 & \textbf{0.970 $\pm$ 0.081} \\
\bottomrule
\end{tabular}
}
\end{table}

Beyond volumetric overlap, we scrutinized surface accuracy using the Normalized Surface Distance (NSD) at a clinically stringent tolerance of 1mm. As shown in Table \ref{tab:segmentation}, the Average NSD significantly improves to $0.970$ ($p = 0.031$). This metric is vital for surgical guide fabrication, where deviations exceeding 1mm can lead to ill-fitting appliances. It is worth noting that in the Mandible, where the baseline NSD was already robust ($0.968$) due to the high contrast of cortical bone, our method maintained high fidelity ($0.971$, $p=0.198$). This lack of divergence is a critical safety indicator: it confirms that PGMP preserves valid anatomical geometry without introducing structural hallucinations that could degrade otherwise healthy regions, addressing a key safety concern in generative restoration.

%% Insertion of Figure 10 (Segmentation)
\begin{figure}[t]
	\centering
	\includegraphics[width=1.0\linewidth]{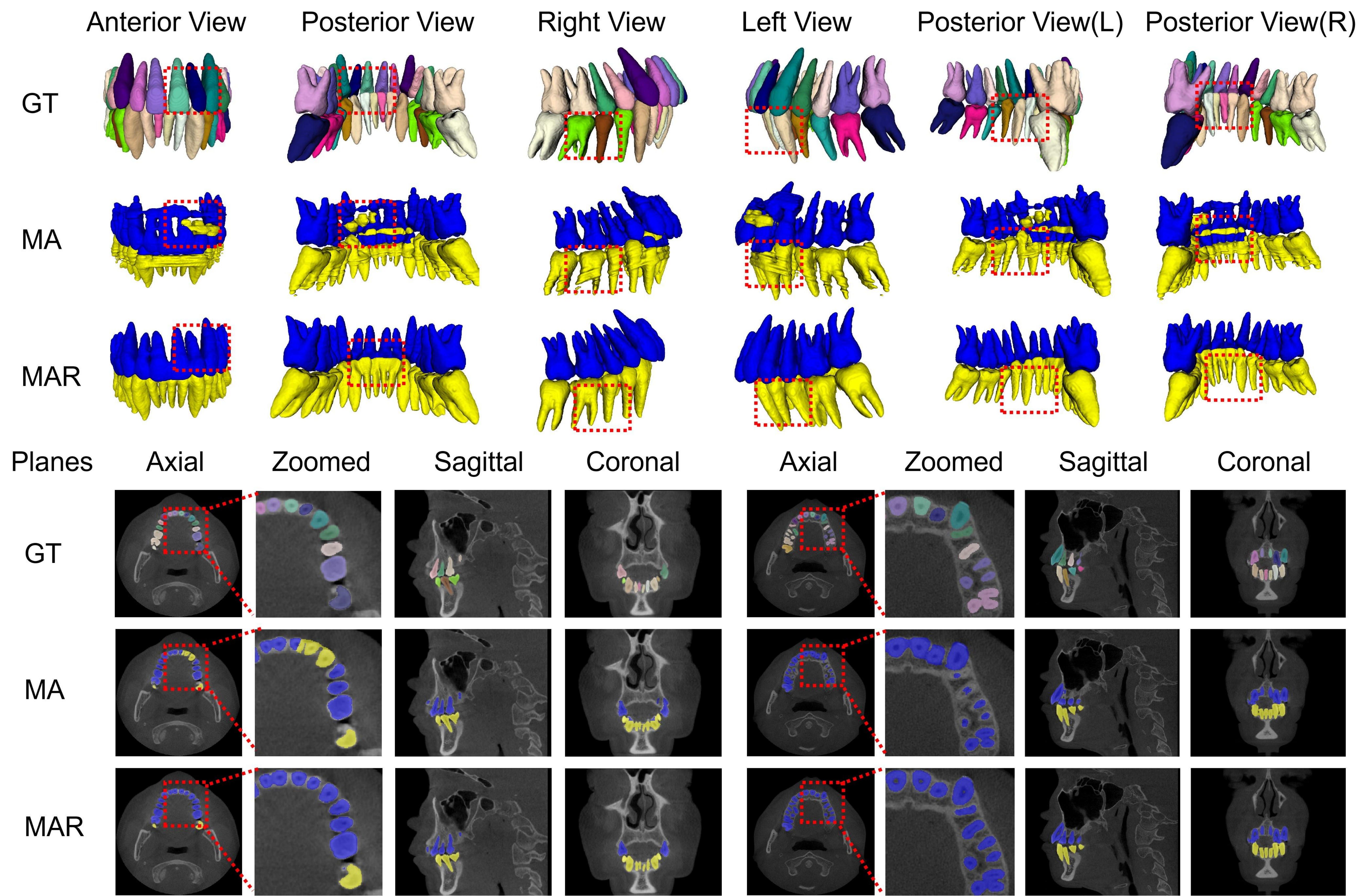}
	\caption{Visual comparison of downstream tooth segmentation results on unseen patient data. GT: Ground Truth segmentation manually annotated by experts. MA: Segmentation on the original metal-artifact image suffers from severe failures, specifically manifesting as significant voids (red boxes) where the metal streaks obscure the root apices, leading to broken dental arches. MAR: Segmentation on the PGMP-restored image successfully recovers the full tooth morphology.  The reconstructed masks (Green) align closely with the Ground Truth, restoring the continuity of the occluded roots and the integrity of the alveolar bone boundaries.}
	\label{fig:segmentation}
\end{figure}

This restoration of geometric fidelity is visually corroborated in Fig. \ref{fig:segmentation}. The segmentation results on the original metal-artifact images reveal critical failures: strong beam-hardening streaks create "shadow zones" interpreted as non-tissue, resulting in missing teeth (red boxes). In contrast, PGMP-restored images exhibit recovered topological continuity, allowing the network to correctly segment the full dental arch, including root apices previously occluded by artifacts. This substantial boost proves that DMP-Former effectively recovers the true geometric fidelity of the hard tissue, establishing itself as a reliable pre-processing module for clinical automation pipelines.

\subsection{Limitations and Future Scope}
\label{sec:limitations}

While the proposed PGMP framework establishes a new benchmark for dental MAR, a transparent analysis of its current boundaries is essential to guide future clinical deployment. A primary constraint lies in the trade-off between physical fidelity and computational feasibility within the AAPS simulation. Although our convolutional scattering proxy effectively captures the low-frequency bias of Compton scattering, it simplifies the complex angular dependencies of photon interactions. Consequently, in complex clinical scenarios involving high-density clusters of multiple implants (e.g., ``all-on-4'' restorations), the model may under-correct high-order scattering interferences, such as dark streaks connecting adjacent screws, which a full Monte Carlo transport simulation would theoretically resolve. Similarly, as the current simulation is calibrated to a standard 120 kVp spectrum, extreme deviations in acquisition protocols—such as low-dose pediatric settings—may necessitate protocol-specific fine-tuning to ensure robust generalization.

Beyond physical approximations, the reliance on foundation model priors introduces a subtle safety paradox regarding semantic interpretation. While MedDINOv3 serves as a powerful ``Digital Anatomist'' for healthy tissue, it inevitably encodes the normative priors of its training distribution. In rare pathological instances, such as large osteolytic cysts or complex jaw fractures adjacent to metal, there exists a theoretical risk that the Semantic-Structural Alignment mechanism might attempt to ``correct'' these lesions into healthy trabecular bone . This ``hallucination of health'' represents a critical safety challenge in generative restoration. Future iterations must therefore incorporate uncertainty-aware anomaly detection mechanisms, allowing the network to automatically down-weight the semantic prior in regions where the input deviates significantly from the normative manifold.

From an architectural perspective, our current implementation operates on 2D slices to leverage the slice-wise attention efficiency of standard Vision Transformers. Although we enforce volumetric consistency during the data generation phase, the restoration network itself lacks explicit 3D attention mechanisms. Consequently, longitudinal continuity along the z-axis relies largely on the inherent smoothness of the anatomy rather than learned inter-slice dependencies. Extending the DMP-Former to pseudo-3D or full-3D architectures remains a promising avenue to further enhance volumetric consistency, provided the quadratic increase in memory consumption can be managed.

%% ===============================================
%% Section 5: Conclusion
%% ===============================================
\section{Conclusion}
\label{sec:conclusion}

In this work, we presented Physically-Grounded Manifold Projection, a framework that strategically adapts the Direct $x$-Prediction paradigm to the challenge of Dental CBCT Metal Artifact Reduction. Motivated by the clinical need to balance restoration fidelity with inference efficiency, our methodology rigorously harmonizes physical laws, geometric constraints, and semantic priors.
The proposed solution functions as a synergistic triad. It begins with an Anatomically-Adaptive Physics Simulation that bridges the synthetic-to-real domain gap via polychromatic spectral modeling, establishing a robust data foundation. Building upon this, the DMP-Former executes a deterministic projection from the artifact space to the anatomical manifold. Crucially, this restoration is guided by a "Digital Anatomist" (MedDINOv3), utilizing medical foundation priors to curb the structural hallucinations that frequently compromise unsupervised approaches.

Extensive empirical validation confirms that PGMP delivers not only superior quantitative metrics but also tangible clinical utility. The strong statistical agreement between automated quality assessments and human radiologist feedback, coupled with statistically significant improvements in downstream segmentation tasks ($p < 0.05$), validates the method's reliability for high-precision digital dentistry workflows. Looking forward, the concept of ``Foundation Model-Guided Manifold Projection'' offers a scalable pathway for medical image restoration. We anticipate extending this philosophy to other inverse problems, such as beam-hardening correction in orthopedic CT or low-dose PET reconstruction, paving the way for the next generation of physics-aware and clinically trustworthy AI systems.

\section*{Acknowledgments}
This research was supported in part by the National Natural Science Foundation of China under Grants No. 62571173 and No. 62206242, and the Zhejiang Provincial Natural Science Foundation of China under Grants No. LD25F020005 and No. LQN25F030009.

% \end{document}

%%Graphical abstract
\begin{graphicalabstract}
\end{graphicalabstract}

%%Research highlights
\begin{highlights}
\item Research highlight 1
\item Research highlight 2
\end{highlights}

\bibliographystyle{elsarticle-num}

%% 指定您的 .bib 文件名 (不需要加 .bib 后缀)
\bibliography{ref}

%% 下面的手动部分应当被删除或注释掉，因为我们现在使用 BibTeX 自动生成
% \begin{thebibliography}{00}
% \bibitem{label}
% Text of bibliographic item
% \end{thebibliography}

\end{document}